\documentclass[a4paper,fleqn]{cas-dc}
\usepackage[font=normalsize]{caption}
\usepackage[authoryear]{natbib}
\usepackage{algorithm}
\usepackage{algorithmic}
\usepackage{textcomp}
\usepackage{url}
\usepackage{verbatim}
\usepackage{svg}
\usepackage{xcolor}

\begin{document}
\let\WriteBookmarks\relax
\def\floatpagepagefraction{1}
\def\textpagefraction{.001}

\shorttitle{}
\shortauthors{Y. Hu et al.}

\title[mode = title]{Global Cross-Modal Geo-Localization: A Million-Scale Dataset and a Physical Consistency Learning Framework}

\author[1]{Yutong Hu}[orcid=0009-0003-1924-4005]
\ead{yutonghu@whu.edu.cn}
\credit{Conceptualization, Methodology, Software, Data curation, Visualization, Writing - original draft}

\author[1]{Jinhui Chen}
\credit{Software, Validation, Formal analysis, Writing - review \& editing}

\author[1]{Chaoqiang Xu}
\credit{Software, Validation, Formal analysis, Writing - review \& editing}

\author[2]{Yuan Kou}
\credit{Resources, Data curation, Writing - review \& editing}

\author[2]{Sili Zhou}
\credit{Resources, Data curation, Writing - review \& editing}

\author[1]{Shaocheng Yan}
\credit{Methodology, Validation, Writing - review \& editing}

\author[1]{Pengcheng Shi}
\credit{Methodology, Supervision, Writing - review \& editing}

\author[1]{Qingwu Hu}
\credit{Resources, Supervision, Funding acquisition, Writing - review \& editing}

\author[1]{Jiayuan Li}[orcid=0000-0002-9850-1668]
\cormark[1]
\ead{ljy_whu_2012@whu.edu.cn}
\credit{Conceptualization, Methodology, Supervision, Funding acquisition, Writing - review \& editing}

\affiliation[1]{organization={School of Remote Sensing and Information Engineering, Wuhan University},
            city={Wuhan},
            postcode={430079},
            country={China}}

\affiliation[2]{organization={First Surveying and Mapping Institute of Hunan Province},
            city={Changsha},
            postcode={421001},
            country={China}}

\cortext[1]{Corresponding author.}

\begin{abstract}
Cross-modal Geo-localization (CMGL) matches ground-level text descriptions with geo-tagged aerial imagery, which is crucial for pedestrian navigation and emergency response. However, existing studies are constrained by narrow geographic coverage and simplistic scene diversity, failing to reflect the immense spatial heterogeneity of global architectural styles and topographic features. To bridge this gap and facilitate universal positioning, we introduce CORE, the first million-scale dataset dedicated to global CMGL. CORE comprises 1,034,786 cross-view images sampled from 225 distinct geographic regions across six continents, offering an unprecedented variety of perspectives in varying environmental conditions and urban layouts. We leverage the zero-shot reasoning of Large Vision-Language Models (LVLMs) to synthesize high-quality scene descriptions rich in discriminative cues. Furthermore, we propose a physical-law-aware network (PLANET) for cross-modal geo-localization. PLANET introduces a novel contrastive learning paradigm to guide textual representations in capturing {\color{black}statistical scene descriptors} of satellite imagery. Extensive experiments across varied geographic regions demonstrate that PLANET significantly outperforms state-of-the-art methods, establishing a new benchmark for robust, global-scale geo-localization. The dataset and source code will be released at https://github.com/YtH0823/CORE. 
\end{abstract}


\maketitle

\section{Introduction}
The advancement of geo-spatial artificial intelligence (GeoAI) is profoundly deepening humanity's understanding of the physical world. As a fundamental task within this domain, geo-localization enables the determination of real-time coordinates for ground observation points in GNSS-denied environments, which is critical for applications such as pedestrian navigation \citep{function1} and robot localization \citep{function2}. Traditional geo-localization methods primarily rely on Cross-View Geo-localization (CVGL). CVGL achieves positioning by performing cross-view retrieval between ground-level imagery and geo-tagged satellite images, a technique that has already seen widespread adoption \citep{Sample4geo,DAC,camp,GeoDTR,transgeo,robustlocation,geodtr+}. However, CVGL necessitates the provision of real-time ground images from the query side. In certain restricted environments, capturing and uploading images can be impractical, thereby limiting the operational scope of such positioning systems. To enhance flexibility, recent studies \citep{geotext1652,CVGText} have begun exploring Cross-Modal Geo-localization (CMGL) based on natural language. CMGL directly matches descriptive text of ground scenes with satellite imagery, offering significant value in real-world scenarios. For instance, taxi drivers frequently determine locations based on passengers' verbal descriptions \citep{text2pos}. Similarly, in emergency calls, individuals seeking help often can only describe their surroundings through language. CMGL leverages this textual information for precise localization, which is of vital importance for improving the efficiency of emergency response \citep{TEXTFUNCTION1}. {\color{black}Thus, CMGL provides a viable way to bridge the localization gap when ground-level images are unavailable.}

Images from diverse geographic regions exhibit significant visual discrepancies that inevitably induce corresponding divergences in textual descriptions, resulting in a spatial heterogeneity that renders worldwide CMGL an exceptionally challenging task. This divergence stems from two primary factors. From a natural perspective, varying climatic conditions across latitudes result in diverse topographic features \citep{motivation1}. From a humanistic perspective, historical and cultural differences lead to distinct architectural styles. To ensure the global generalization of the model, learning from diverse global imagery is essential \citep{motivation2}. Furthermore, to achieve fine-grained visual-semantic alignment, the model must be trained on exhaustive descriptions of ground scenes, which imposes strict standards on both datasets and methodologies. Although some studies have attempted to promote global-scale localization \citep{CVGText,geobridge}, there remain deficiencies in data volume and modal diversity.

\begin{figure*}[pos=t]
\centering
\includegraphics[width=6.4 in]{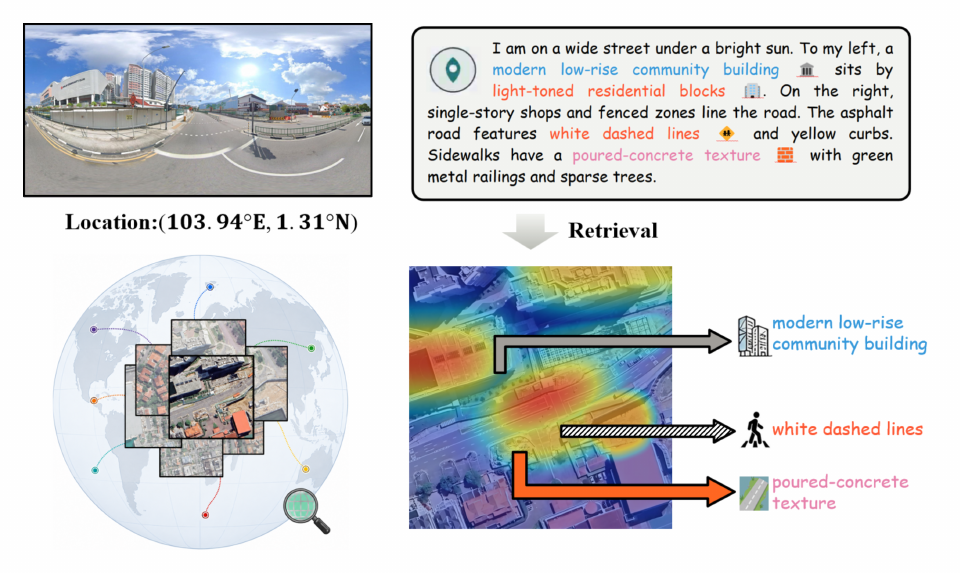}
\caption{Illustration of cross-modal geo-localization. This task retrieves satellite imagery based on text queries of the ground environment for position localization. The physical attribute information presented in the text description can yield valuable clues regarding the location.}
\label{TEASER}
\end{figure*}

\begin{table*}[pos=t]
\centering
\caption{Comparison of the CORE dataset with other geo-localization datasets. The comparison includes the number of reference and query images, text annotations, distribution range, and sampling environment.}
\label{tab:dataset_comparison}

\resizebox{\textwidth}{!}{%
\begin{tabular}{lccccccc}
\toprule
\textbf{Dataset} & {\color{black}\textbf{Image Type}} & \textbf{Query Image} & \textbf{Reference Image} & \textbf{Total Image} & \textbf{Text Annotation} & \textbf{Distribution} & \textbf{Environment} \\
\midrule
\textbf{CVGL} & & & & & & & \\
CVUSA \citep{CVUSA} & {\color{black}Street-view} & 44,416 & 44,416 & 88,832 & $\times$ & USA & Suburban \\
CVACT \citep{CVACT} & {\color{black}Street-view} & 128,334 & 128,334 & 256,668 & $\times$ & Canberra & Urban \\
VIGOR \citep{vigor} & {\color{black}Street-view} & 90,618 & 105,214 & 195,832 & $\times$ & USA & Urban \\
University-1652 \citep{U1652} & {\color{black}Drone and Street-view} & 50,218 & 55,227 & 105,445 & $\times$ & Campus & Urban \\
CVGlobal \citep{CVGlobal} & {\color{black}Street-view} & 134,233 & 134,233 & 268,466 & $\times$ & Global (7 regions) & Urban \\
CV-Cities \citep{CVCITIES} & {\color{black}Street-view} & 223,736 & 223,736 & 447,472 & $\times$ & Global (16 regions) & Urban \\
DReSS \citep{dress} & {\color{black}Street-view} & 422,760 & 174,934 & 597,694 & $\times$ & Global (8 regions) & Urban and Suburban \\
\midrule
\textbf{CMGL} & & & & & & & \\
Geotext-1652 \citep{geotext1652} & {\color{black}Drone and Street-view} & 50,218 & 55,227 & 105,445 & \checkmark & Campus & Urban \\
CVG-Text \citep{CVGText} & {\color{black}Street-view} & 39,792 & 79,584 & 119,376 & \checkmark & 3 Cities & Urban and Suburban \\
GeoLoc \citep{geobridge} & {\color{black}Drone and Street-view} & 52,684 & 105,362 & 158,046 & \checkmark & Global (36 regions) & Urban and Suburban \\
\midrule
\textbf{CORE (Ours)} & {\color{black}\textbf{Street-view}} & \textbf{517,393} & \textbf{517,393} & \textbf{1,034,786} & \checkmark & \textbf{Global (225 regions)} & Urban and Suburban \\
\bottomrule
\end{tabular}
}
\vspace{1pt} 
\begin{flushleft}
\footnotesize *Note: CVG-Text covers New York, Brisbane, and Tokyo.
\end{flushleft}
\end{table*}

As shown in Table~\ref{tab:dataset_comparison}, existing popular datasets \citep{CVUSA,CVACT,vigor,CVGlobal,U1652,CVCITIES,dress,geotext1652,CVGText,geobridge} are flawed in scale and geographic coverage. The lack of text annotations \citep{CVUSA,CVACT,vigor,CVGlobal,U1652,CVCITIES,dress} also prevents them from supporting CMGL tasks. Although some datasets \citep{geotext1652,CVGText,geobridge} include textual descriptions, their data collection is limited. To address this scarcity, we introduce \textbf{C}ross-modal ge\textbf{O}-localization across wo\textbf{R}ldwide r\textbf{E}gions (\textbf{CORE}), the first million-scale global CMGL dataset. {\color{black}CORE is positioned as a large-scale worldwide benchmark for text-to-satellite geo-localization based on ground-level observations.} {\color{black}It contains 517,393 street-satellite pairs} from 225 diverse locations globally. Each ground-level image is paired with a fine-grained textual annotation. {\color{black}Guided by annotation rules summarized from human descriptions, the text captures clear location cues and spatial relations in the ground scene, reducing ambiguity and supporting fine-grained alignment. Because the text is derived from ground-level observations, its semantics naturally differ from those visible in overhead satellite imagery. CORE intentionally retains this cross-view semantic gap, making it a meaningful benchmark for evaluating the inherent challenge of CMGL. The global sampling further introduces region-dependent language-visual variation caused by differences in architectural styles and environmental conditions, enabling CORE to evaluate CMGL models under worldwide geographic heterogeneity.} Compared with existing datasets, CORE stands out through its vast geographic diversity. This advantage allows it to better meet the requirements for complex scene perception in global navigation. In short, CORE offers comprehensive advantages and notable merits over existing geo-localization datasets.

The essence of CMGL lies in the cross-modal retrieval between ground descriptions and aerial imagery. Its primary goal is to establish semantic associations within a unified latent space. However, existing image-text retrieval methods \citep{CLIP,blip,remoteclip,longclip,EVAClip} predominantly rely on traditional contrastive learning paradigms, which only optimize the global similarity between feature vectors \citep{contrastivelearning}. While these methods can align macroscopic semantics, they struggle to capture the inherent fine-grained physical attributes within images through coarse-grained global matching. As illustrated in Fig.~\ref{TEASER}, physical attributes such as color, spatial structure, and texture are regularly reflected in the pixel distribution of images and are transformed into semantic cues embedded in the text during scene description. To establish a fine-grained mapping between textual semantics and underlying image attributes, we propose a physical-law-aware network (PLANET). Specifically, PLANET introduces a novel contrastive learning paradigm, which utilizes physical semantic projectors to map textual representations into fine-grained physical attribute descriptors. On the image side, {\color{black}in addition to global deep features,} we extract a series of deterministic, parameter-free {\color{black}statistical descriptors} by decoupling the statistical distribution patterns of the images \citep{experience1,experience2}. These classic heuristic signatures explicitly characterize surface morphology and optical features, serving as {\color{black}complementary statistical cues for training}. By explicitly associating textual physical attribute descriptors with {\color{black}these statistical descriptors}, text branches can be effectively driven to learn and model geographical prior constraints in different environments. This alignment mechanism endows textual features with a high degree of environmental-discriminative power consistent with images, promoting stronger convergence of cross-modal representations within the same environment in the manifold space. By enhancing the feature's ability to perceive environmental heterogeneity, PLANET fundamentally improves its localization robustness in complex scenes.

\begin{figure}[pos=t]
\centering
\makebox[\columnwidth][c]{\includegraphics[width=1.12\columnwidth,trim=6 6 6 6,clip]{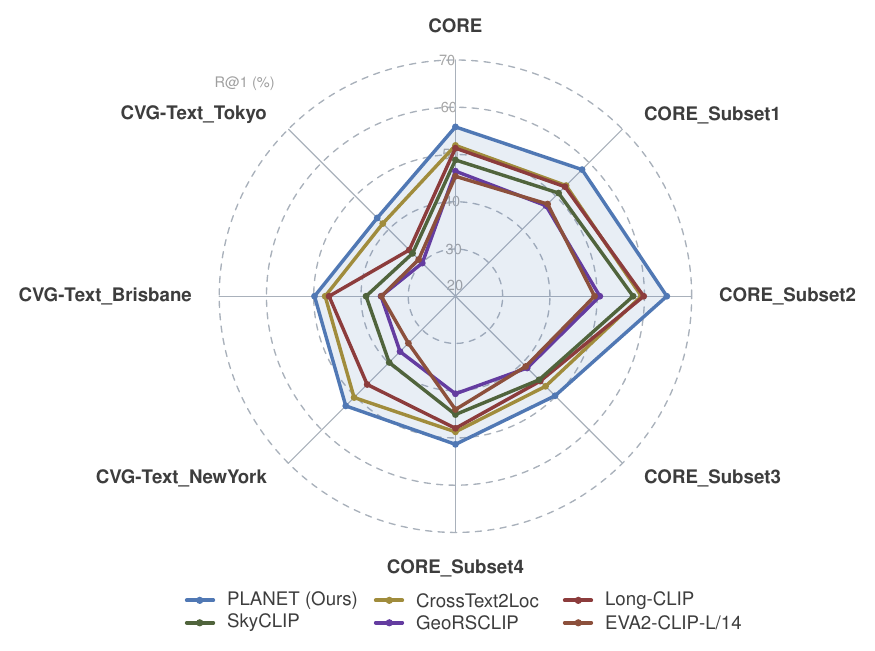}}
\caption{We evaluate representative methods on the CORE dataset and the existing CVG-Text dataset, with PLANET achieving state-of-the-art performance in 8 settings, demonstrating strong localization capabilities. All results are reported as R@1, {\color{black}and the radial axis is linearly scaled from 20 to 70.}}
\label{RESULT}
\end{figure}

We conduct extensive experiments across diverse datasets and geographical settings. The quantitative results presented in Fig.~\ref{RESULT} indicate that our proposed PLANET outperforms established methods \citep{CLIP,CVGText,blip,tagclip,longclip} across various regions of the CORE dataset and other datasets. In summary, the CORE dataset and the PLANET framework aim to establish a large-scale benchmark for CMGL on a global scale. This benchmark will facilitate the development of new algorithms. Our contributions can be summarized as follows:

(1) We propose CORE, the first million-scale dataset for the global-scale CMGL task. Encompassing 1,034,786 multi-view images with corresponding fine-grained text annotations across 225 representative geographic regions, CORE bridges existing data gaps through its diverse scene representations.

(2) A physical-law-aware network (i.e., PLANET) is proposed to address the heterogeneity of geographical distribution. By introducing a consistency contrastive learning paradigm, this network can accurately mine potential semantic cues in text and explicitly align them with {\color{black}statistical attributes of images}, significantly enhancing the model's feature discrimination ability.

(3) Utilizing the CORE dataset, we construct a worldwide CMGL benchmark to assess large-scale geo-localization under geographic variability. Experimental results indicate that PLANET achieves state-of-the-art performance on this benchmark. Furthermore, we demonstrate the benchmark's effectiveness in evaluating model adaptation to shifts in geographic distribution through cross-domain generalization experiments.

\begin{figure*}[pos=t]
\centering
\includegraphics[width=7.0 in]{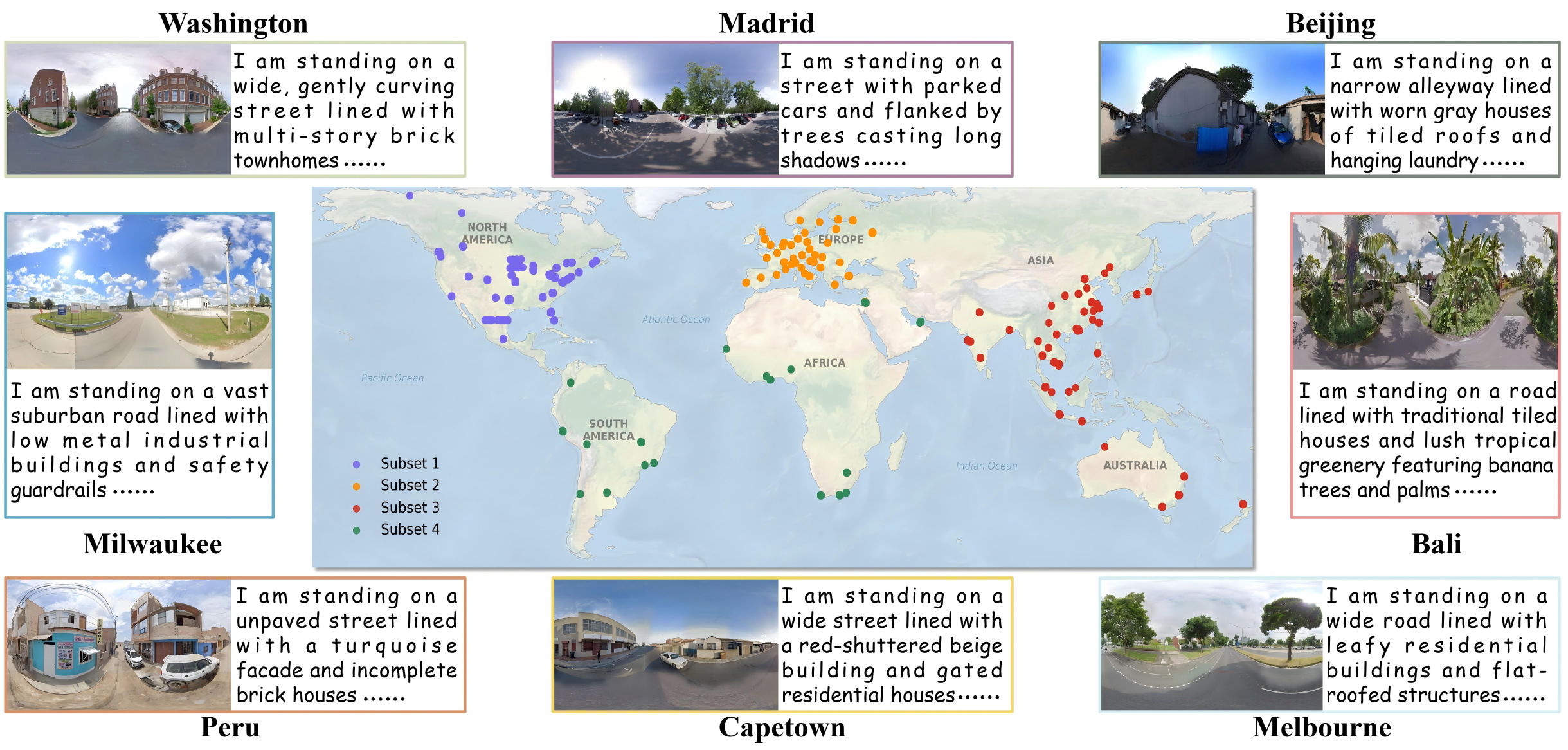}
\caption{Geographical distribution map of sampled images and corresponding text annotations from the CORE dataset. CORE exhibits unprecedented global scale and spatial heterogeneity. This extensive diversity ensures a comprehensive representation of varying urban and suburban environments worldwide.}
\label{fig3}
\end{figure*}

\section{Related Work}
In this section, we first discuss existing datasets for geo-localization. Subsequently, we review current localization methods and potential methods derived from the field of text-image retrieval.
\subsection{Geo-localization Datasets}
As illustrated in Table~\ref{tab:dataset_comparison}, geo-localization datasets can be categorized into cross-view and cross-modal geo-localization datasets based on the modality of the query data (e.g., images or text) used for retrieval.

\subsubsection{CVGL Datasets}
Over the past decade, a multitude of datasets have been curated to advance the field of cross-view geo-localization. CVUSA \citep{CVUSA} initially established a large-scale repository of ground and aerial imagery across the United States; its widely adopted version is a refined subset consisting of 35,532 training and 8,884 testing pairs. Following a similar paradigm, CVACT \citep{CVACT} provides extensive data splits, offering robust empirical support for model evaluation. The VIGOR dataset \citep{vigor}, comprising 105,214 ground images and 90,618 aerial images, further advances the field by relaxing the rigid center-alignment assumption, allowing query points to be stochastically distributed within target regions.

Although these datasets successfully simulate realistic scenarios, their data collection is confined to a single nation, which inherently limits the model's ability to generalize across the vast geographic domain gaps arising from variations in climate, vegetation, and architectural styles. While CVGlobal \citep{CVGlobal}, CV-Cities \citep{CVCITIES}, and DReSS \citep{dress} attempt to broaden the scope by sampling across multiple international cities, the inherent spatial sparsity and discontinuous distribution of their data points fail to capture the statistical complexity of global geographic features, thereby falling short of supporting true worldwide geo-localization. University-1652 \citep{U1652} provides imagery covering 1,652 groups of satellite and simulated drone images across 72 university campuses worldwide. However, a single campus environment is too limited to support the requirements for large-scale geo-localization. In summary, existing cross-view datasets are insufficient in both data scale and geographic diversity to support true worldwide geo-localization. Furthermore, the reliance on a single visual modality fails to meet the practical multi-modal interaction requirements of pedestrians, tourists, and other end-users.

\subsubsection{CMGL Datasets}
The evolution of Large Language Models (LLMs) has made it feasible to equip ground-level imagery with precise textual annotations \citep{LLMLABEL1,LLMLABEL2,llava,RSICD}. To address specific scenarios with high demands for natural language interaction, such as urban navigation \citep{urbannavigation} and emergency response \citep{disasterresponse}, researchers have developed cross-modal geo-localization datasets pairing text with satellite imagery. However, as illustrated in Table~\ref{tab:dataset_comparison}, only three publicly available datasets are specifically designed for the cross-modal geo-localization task, and their limited scale and regional coverage still pose a significant constraint on the advancement of this field.

\textit{Geotext-1652 Dataset}: Chu et al. \citep{geotext1652} proposed the pioneering CMGL dataset tailored for drone navigation. By augmenting UAV images from the image-based University-1652 with global textual descriptions, this dataset effectively extends the benchmark into the cross-modal domain. However, the dataset has certain limitations. Due to the constrained coverage of the synthetic drone imagery, the textual annotations struggle to reach a semantic granularity sufficient for characterizing scene details, resulting in low retrieval accuracy.

\textit{CVG-Text Dataset}: Ye et al. \citep{CVGText} introduced a city-level CMGL dataset, employing GPT-4o \citep{gpt} to generate descriptive text for 30,000 scene points across three major cities, i.e., New York, Brisbane and Tokyo. While this dataset enhances textual richness through Large Language Models, its restricted geographical span limits the model's capability to represent complex natural landscapes. Furthermore, CVG-Text focuses primarily on modern metropolitan environments and generic semantics, leading to insufficient characterization of intrinsic physical attributes and hindering its adaptability to global-scale environmental heterogeneity.

\textit{GeoLoc Dataset}: Recently, Song et al. \citep{geobridge} proposed a CMGL dataset spanning 36 global regions, which encompasses 52,684 multi-view image-text pairs. A notable advantage of this dataset is its support for highly flexible mutual querying between textual descriptions and multi-view imagery. However, constrained by its limited data volume and regional diversity, GeoLoc struggles to comprehensively capture the extensive spatial heterogeneity at a global scale.

Overall, existing dedicated datasets for cross-modal geo-localization remain insufficient to address localization challenges in diverse global environments. Consequently, it is imperative to establish a comprehensive and large-scale CMGL dataset integrating geographically diverse imagery and fine-grained textual descriptions.

\subsection{Geo-localization Methods}
To motivate cross-modal geo-localization on a global scale, we discuss existing CVGL methods in the remote sensing domain and potential techniques from the field of text-to-image retrieval that could be adapted for CMGL in the subsequent sections.

\subsubsection{CVGL Methods}
Cross-View Geo-localization (CVGL) has emerged as a pivotal research area. Workman et al. \citep{CVUSA} pioneered this field by demonstrating that deep neural features offer superior generalization over traditional hand-crafted descriptors. Early research primarily focused on establishing robust feature extraction frameworks \citep{CNN1,CNN2,CVMNET}, while contemporary approaches emphasize architectural modernization by integrating Transformer-based global modeling and MLP-Mixer designs \citep{transformer1,transgeo,camp,DAC,CVCITIES} to capture long-range spatial dependencies. {\color{black}Recent CVGL studies further improve localization robustness by enhancing cross-view feature interaction and spatial alignment \NoHyper\citep{DRL,p2fcn,EAGLe}\endNoHyper.} Beyond feature extraction, a fundamental motivation in CVGL is to bridge the geometric and semantic discrepancies between disparate perspectives. {\color{black}Recent studies on cross-view representation learning have explored view-aware semantic alignment \NoHyper\citep{Zhang_2026_CVPR}\endNoHyper{} and representation decoupling \NoHyper\citep{zhang2024cvpr}\endNoHyper{} to preserve discriminative information under substantial viewpoint changes.} Some methodologies strive to achieve cross-view alignment through contrastive learning and hard negative mining, thereby enhancing the discriminative power of joint embeddings \citep{Sample4geo,hardsample}. Others introduce geometric transformations, such as Bird’s-Eye-View (BEV) projections, to explicitly rectify perspective distortions between street-level and satellite imagery \citep{CVGlobal,BEV}. Additionally, spatial priors have been incorporated as auxiliary constraints to mitigate the challenges posed by non-center-aligned queries \citep{dress}. {\color{black}Related representation learning methods further improve robustness to structural variations through explicit layout modeling \NoHyper\citep{zhang2022cvpr}\endNoHyper{}, spatial-temporal relational reasoning \NoHyper\citep{zhang2024tpami}\endNoHyper{}, and uncertainty modeling \NoHyper\citep{zhang2024ijcv}\endNoHyper{}.} {\color{black}Despite these advances, CVGL is still built on image-to-image retrieval and generally requires a ground-level query image. When such images are unavailable, text-based queries provide another practical way to support geo-localization, motivating the study of CMGL.}

\subsubsection{Potential Technologies in Text-to-Image Retrieval}

In parallel, the rapid evolution of Vision-Language Models (VLMs) has provided significant technical underpinnings for text-to-image retrieval tasks. Inspired by the success of CLIP \citep{CLIP}, contemporary research has progressed from basic cross-modal alignment to more sophisticated contrastive learning paradigms. For instance, BLIP \citep{blip} and its successors \citep{blip2,blip3} introduce a hybrid encoder-decoder architecture to unify vision-language understanding and generation, providing a robust framework for capturing complex semantic correlations. SigLIP \citep{Siglip} further optimized the learning objective by replacing the conventional softmax-based contrastive loss with a pairwise sigmoid loss, markedly improving training efficiency and stability in large-scale retrieval scenarios. Furthermore, the integration of Large Language Models (LLMs) with visual encoders, as exemplified by ALIGN \citep{Align} and subsequent models \citep{remoteclip,EVAClip,XVLM,X2VLM}, has demonstrated exceptional zero-shot generalization capabilities across diverse domains. Although these methodologies exhibit outstanding performance in general text-to-image matching, their direct migration to CMGL remains challenging. It is noted that standard models are primarily pre-trained on natural images, which lack the specialized geomorphic features and top-down spatial perspectives inherent in remote sensing data. This domain gap, coupled with the absence of fine-grained physical attribute characterization, limits their efficacy for precise localization in heterogeneous global environments. Therefore, customizing text-to-image retrieval architectures for large-scale geographic contexts is of paramount importance.

\section{CORE Dataset}
\label{sec:dataset}
Our objectives for developing a new CMGL dataset are twofold: (i) to fill the gap in large-scale datasets for training highly generalizable models and (ii) to facilitate a meaningful yet formidable new task: worldwide cross-modal geo-localization. This section provides a comprehensive overview of the CORE dataset, focusing on three critical dimensions: data collection, data annotation, and data analysis.

\subsection{Data Collection}
To foster scale-invariant feature learning and enhance out-of-distribution generalization across heterogeneous environments, we curate the CORE dataset by integrating multi-source street-level and satellite imagery. Specifically, {\color{black}we first collect a candidate pool of panoramic street-view images} from existing datasets and high-resolution captures from Google Street View \footnote{https://www.google.com/streetview/} and Baidu Maps \footnote{https://map.baidu.com}. To establish cross-modal pairs, we leveraged the Google Maps API \footnote{https://www.google.com/maps} to retrieve corresponding satellite imagery for each street-level location, covering a local area of 500 $m^2$. {\color{black}After image screening, cross-view pair validation, and annotation review, 517,393 valid street-view--satellite pairs are retained in the final CORE dataset.} These images span 225 distinct geographic regions across six continents, ensuring comprehensive global representation. Each street-level panorama features a resolution of 2,048 $\times$ 1,024 pixels. The spatial resolution of these satellite images ranges from 0.1 m to 0.6 m, providing fine-grained geomorphic details. The global distribution of the dataset and representative samples are illustrated in Fig.~\ref{fig3}.

The collected image pairs are partitioned into four balanced subsets based on the dual criteria of geospatial proximity and geomorphic characteristics, predominantly comprising North America (Subset1), Europe (Subset2), Asia-Oceania (Subset3), and Africa-South America (Subset4). This structured division is primarily designed to mitigate regional biases during model training, preventing the network from overfitting to environmental features or architectural styles unique to a specific continent. Furthermore, such a partitioning scheme facilitates rigorous cross-continent generalization testing, enabling a comprehensive assessment of the model's ability to adapt to heterogeneous and previously unseen global domains.

\subsection{Data Annotation}

\begin{figure}
\centering
\includegraphics[width=\columnwidth]{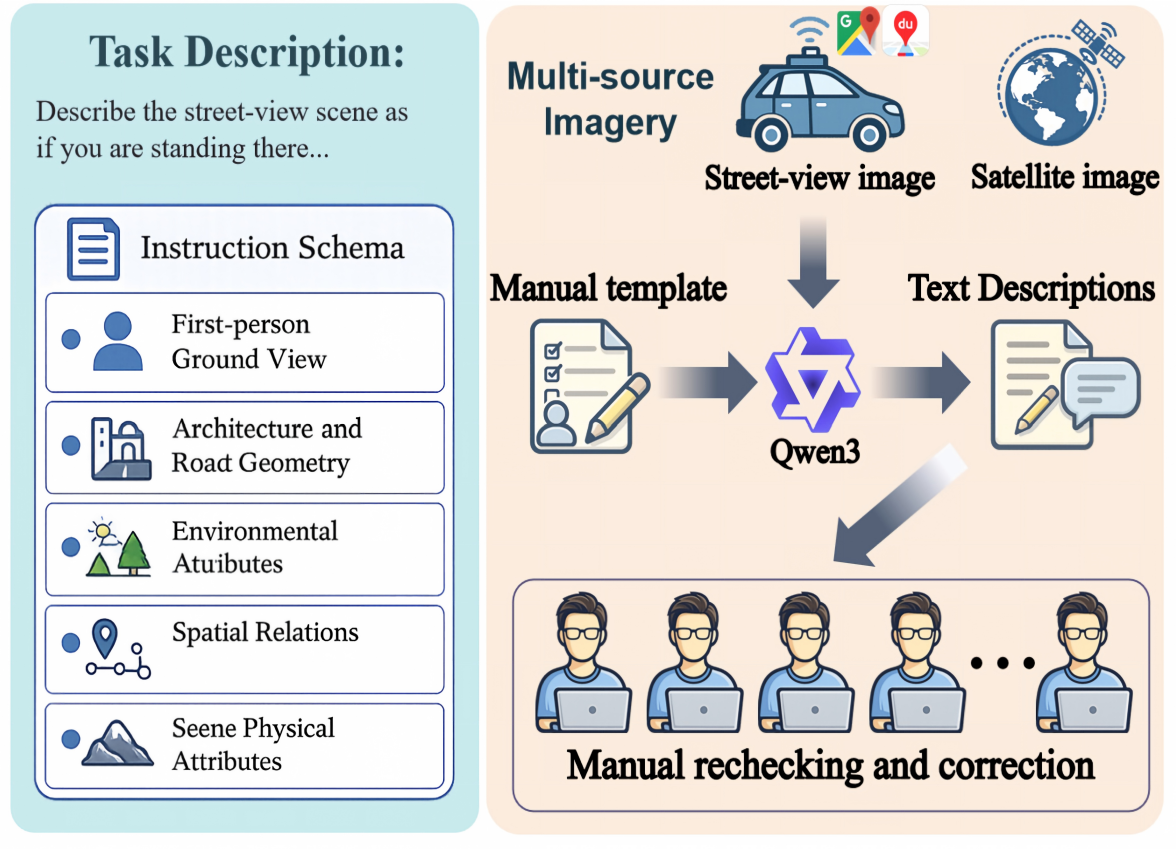}
\caption{Annotation pipeline of the CORE dataset. {\color{black}Manual templates guide Qwen3-VL-Plus generation, and every generated description is manually reviewed and corrected.}}
\label{fig4}
\end{figure}

To mitigate the cross-modal misalignment between visual scenes and natural language, we establish a human-guided prompting framework to generate fine-grained annotations via Qwen3-VL-Plus \footnote{https://github.com/QwenLM/Qwen3-VL}. Fig.~\ref{fig4} illustrates the descriptive process for the CORE dataset.

\noindent Human observers with domain knowledge first manually describe 10\% of the street-view images (approximately 50,000), proportionally sampled from all 225 regions, requiring around 300 hours of work. The human-written descriptions are used to summarize the description logic and construct structured prompts for large-scale annotation. The prompts adopt a first-person pedestrian perspective and describe architectural types, road geometry, environmental elements, physical attributes such as color, material, and vegetation density, and egocentric spatial relations in a unified order. The generated descriptions for the remaining samples are then checked and corrected according to malformed responses, hallucinated landmarks, visual-textual contradictions, or illogical spatial narratives. {\color{black}Descriptions with unsupported objects or landmarks, conflicts with the street-view image, or inconsistent spatial relations are treated as unqualified samples. Such samples are manually corrected when reliable revision is possible and otherwise discarded. More than 90\% of the generated descriptions pass this manual check, and the final 517,393 image-text pairs in CORE do not include unqualified annotations. This human-in-the-loop workflow produces queries with clearer localization cues and richer spatial relations than the direct LVLM-based captioning used in existing CMGL datasets.}

\begin{figure}
\centering
\includegraphics[width=\columnwidth]{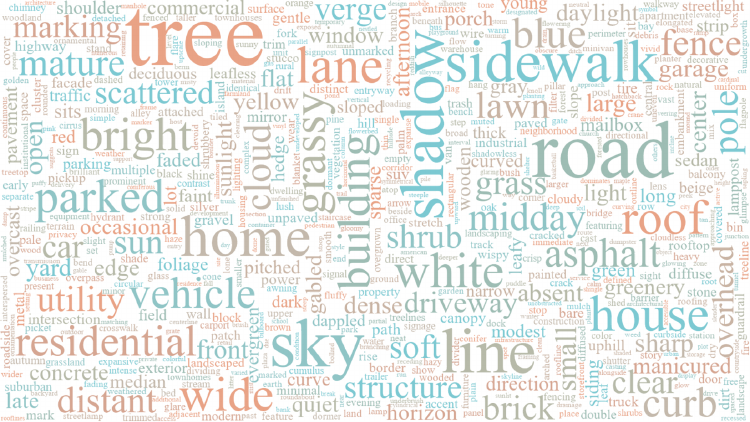}
\caption{Word cloud analysis of text descriptions from the CORE dataset.}
\label{wordcloud}
\end{figure}

\begin{figure*}[pos=t]
\centering
\includegraphics[width=\textwidth]{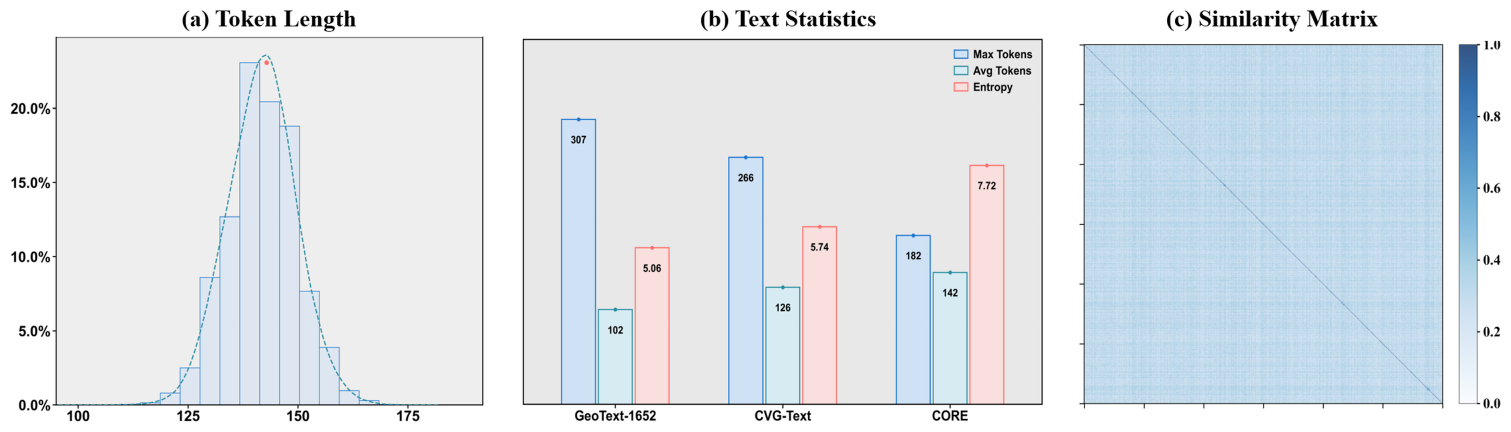}
\caption{Overview of text feature statistics in the CORE dataset. (a) Token length distribution histogram; (b) Text statistics comparison between the CORE dataset and existing CMGL datasets; (c) Text similarity matrix.}
\label{fig6}
\end{figure*}

\subsection{Data Analysis}
Compared with existing datasets in geo-localization, the CORE dataset exhibits unique properties in geographical diversity, scale diversity, and scene diversity. Particular physical attribute mining and fine-grained information density are also defining characteristics that set CORE apart from traditional benchmarks.

\subsubsection{Data Characteristics}
The CORE dataset is distinguished by five characteristics that facilitate robust global-scale geo-localization:

\begin{itemize}
    \item \textbf{Geographical Diversity}: The CORE dataset spans 225 distinct regions across six continents, encompassing a broad spectrum from high-density urban centers to sprawling suburban landscapes. As illustrated in Fig.~\ref{fig3}, this extensive sampling strategy is designed to mitigate overfitting to specific architectural styles or regional surface textures. By introducing high geographical heterogeneity, the dataset encourages models to learn robust, domain-agnostic feature representations. {\color{black}Compared with existing CMGL datasets, this broader sampling provides a more balanced setting for evaluating scene diversity and geographic bias.}
    
    \item \textbf{Scale Diversity}: By integrating multi-source street-level and high-resolution satellite imagery, CORE establishes a hierarchical framework supporting localization at city, continental, and global scales. This multi-scale architecture facilitates fine-grained street-texture matching while enabling cross-regional validation across vast territories. Such a design provides a comprehensive benchmark for evaluating the spatial robustness of localization algorithms at varying granularities.
    
    \item \textbf{Scene Diversity}: To ensure global localization consistency, CORE incorporates a highly heterogeneous scene distribution. As illustrated in Fig.~\ref{fig3}, the global sampling spans multiple continents, achieving multi-dimensional coverage from dense urban centers to expansive suburban areas. This geographic breadth is highly consistent with the fine-grained vocabulary shown in Fig.~\ref{wordcloud}, characterizing both structured urban elements and suburban features such as vegetation, terrain, and road textures. Such coherent modeling, bridging macro-distribution with micro-attributes, enables the data to transcend transient visual appearances for deep semantic alignment.
    
    \item \textbf{Physical Attribute Profiling}: A defining merit of the CORE dataset lies in its emphasis on {\color{black}stable scene appearance cues rather than transient phenomena}. Annotations systematically encode geometric dimensions, material compositions, and chromatic attributes, translating visual perception into structured features. This rigorous descriptive methodology empowers models to achieve a deeper understanding of {\color{black}scene-level environmental characteristics} across heterogeneous geographic scenes.
    
    \item \textbf{Fine-grained Tokens}: Fig.~\ref{fig6}(a) shows that token lengths in CORE primarily range from 125 to 175, peaking at 140. This Gaussian-like, fine-grained sample structure ensures semantic depth while enhancing training robustness. Serving as the query-side, these high-density tokens provide precise signals that enrich the discriminative power of the feature space. Such detailed cues strengthen cross-modal alignment in the manifold space, ensuring superior localization stability even in image-constrained scenarios.
\end{itemize}

\subsubsection{Data Statistics and Visualization}
\label{method}
The quantitative statistical results regarding the textual descriptions of the CORE dataset are illustrated in Fig.~\ref{fig6}(b). The CORE dataset significantly outperforms datasets such as Geotext-1652 and CVG-Text in both average token length and entropy, reflecting superior informational density and scene-characterization capabilities. Meanwhile, the lower maximum token count of CORE effectively circumvents attention dispersion inherent in excessively long sequences. Fig.~\ref{fig6}(c) illustrates a text similarity matrix with minimal cross-sample correlation, verifying the high independence of the generated descriptions. This low-similarity characteristic ensures that each description uniquely and precisely represents its specific geographic scene, effectively mitigating the risk of inter-scene semantic confusion. {\color{black}These statistics provide a quantitative comparison of textual diversity and discriminability with existing CMGL datasets, indicating that the proposed annotation strategy produces richer, less redundant, and more informative queries rather than merely increasing the number of samples.}

\section{Method}

\begin{figure*}[pos=t]
\centering
\includegraphics[width=6.8 in]{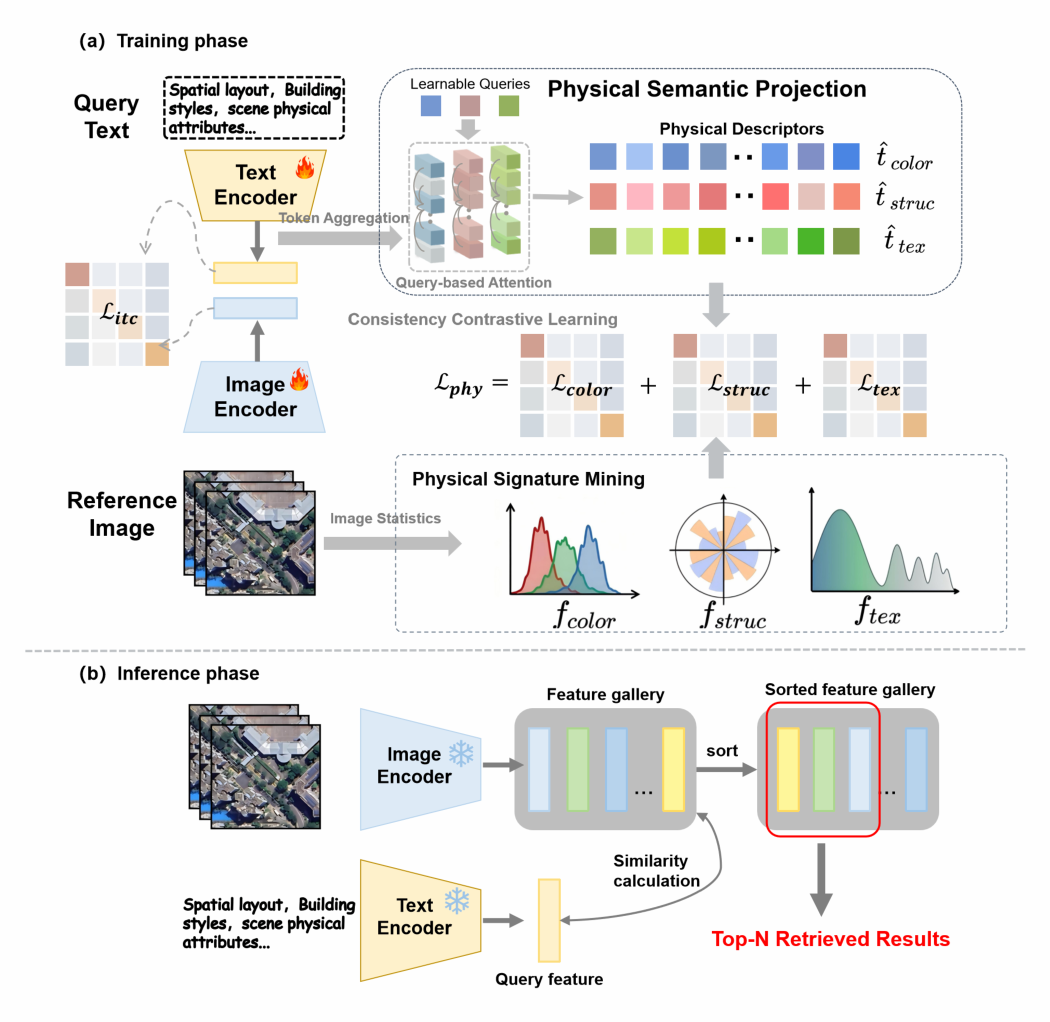}
\caption{Pipeline of the proposed PLANET. {\color{black}(a) During training, image statistics are computed directly from the input satellite images and aligned with the projected text descriptors. (b) During inference, global query and gallery features are matched by similarity without the auxiliary statistical branches.}}
\label{pipeline}
\end{figure*}

Formally, CMGL is formulated as a fine-grained image-text retrieval task. Given a query set of street-view descriptions $\mathcal{T}$ and a gallery of $N$ geo-tagged satellite images $\mathcal{S} = \{I_1, I_2, \dots, I_N\}$, the objective is to retrieve the ground-truth image spatially corresponding to a specific query $T \in \mathcal{T}$. This process is mathematically defined as identifying the target image $\hat{I}$ that maximizes the cross-modal similarity in a shared latent space:

\begin{equation}
\hat{I} = \operatorname*{arg\,max}_{I_k \in \mathcal{S}} \operatorname{sim}(\Phi_{txt}(T), \Phi_{img}(I_k))
\end{equation}

\noindent where $\Phi_{txt}(\cdot)$ and $\Phi_{img}(\cdot)$ denote the textual and visual encoders that extract global semantic representations, and $\operatorname{sim}(\cdot, \cdot)$ represents the cosine similarity metric. Building upon this general retrieval paradigm, the proposed PLANET framework introduces {\color{black}statistical constraints at the scene level} to enhance the environmental discriminability of the feature encoders.

Fig.~\ref{pipeline} presents the overall architecture of the proposed PLANET, which is composed of three integral parts. First, considering that standard encoders struggle to capture inherent fine-grained physical attributes from pixel distributions, an intrinsic physical signature mining module is proposed. This module explicitly decouples the statistical distribution patterns of satellite imagery to mine {\color{black}statistical descriptors of satellite scene appearance}, accurately characterizing surface morphology across color, spatial structure, and texture dimensions. Second, for the main backbone, the framework adopts a general dual-stream vision-language architecture, employing a visual encoder and a textual encoder to independently extract global semantic representations from satellite imagery and street-view descriptions, respectively. To bridge the semantic gap between abstract linguistic concepts and visual appearance, physical semantic projectors are introduced following the textual branch. These projectors map high-level textual features into {\color{black}fine-grained attribute descriptors}, effectively translating linguistic descriptions into {\color{black}structured attribute representations} while maintaining global feature extraction. Finally, to ensure precise cross-modal interaction at the attribute level, a consistency contrastive learning paradigm is introduced. This mechanism enforces the alignment between {\color{black}the statistical descriptors mined from images} and the projected textual descriptors, guiding the textual branch to model geographical prior constraints. By jointly optimizing the global contrastive loss and this physical consistency constraint, the model's environmental-discriminative power in heterogeneous global scenarios is significantly enhanced.

\subsection{Intrinsic Physical Signature Mining}

\label{sec:mining}

{\color{black}Global-scale CMGL faces substantial geographic heterogeneity and cross-view semantic discrepancies, making pure global contrastive learning prone to coarse semantic alignment while overlooking fine-grained localization cues.} To endow the model with explicit perception of the physical environment, we design the \textbf{Intrinsic Physical Signature Mining} module {\color{black}to extract deterministic statistical distributions from satellite imagery as auxiliary supervision, thereby guiding the text branch toward scene cues related to physical appearance.}

Distinct from traditional methods \citep{handfeature,handfeature2}, which treat handcrafted features as supplementary inputs to the network, our motivation is fundamentally different: \textbf{we utilize these statistical characteristics as supervision signals, not input features.} {\color{black}In PLANET, these deterministic statistics form an interpretable scene-level supervision bridge between satellite appearance and textual representations, rather than serving as strict laws or exact labels.} By mining intrinsic statistical distribution patterns from satellite imagery, we construct a set of {\color{black}statistical descriptors without learnable parameters} designed to guide the deep network in learning a semantic space {\color{black}that better reflects satellite scene appearance}, rather than merely unexplainable features. This constitutes a physical consistency mechanism. Specifically, by explicitly decoupling the statistical distribution patterns of the images, we extract {\color{black}interpretable statistical signatures} along {\color{black}three physically motivated dimensions}: spectral chromaticity, geometric structure, and surface texture.

\subsubsection{Spectral Chromaticity Mining}
{\color{black}Scene-level color distribution provides an intuitive statistical cue} for distinguishing land cover types (e.g., vegetation, water bodies, man-made structures). To capture macro-level spectral priors that reflect the statistical expectation of material composition, we utilize statistical color histograms \citep{color} as supervision targets. Given an input image $I$, each color channel is quantized into fixed discrete bins. The normalized frequency $h_{color}^{c}(k)$ for the $k$-th bin of channel $c$ is calculated as:
\begin{equation}
    h_{color}^{c}(k)=\frac{1}{HW}\sum_{x,y}\mathbb{I}(I_{c}(x,y)\in bin_{k})
    \label{eq:color_hist}
\end{equation}
where $\mathbb{I}(\cdot)$ is the indicator function. The concatenation of histograms from all channels constitutes the spectral chromaticity signature $f_{color}$, providing prior constraints regarding material composition. {\color{black}As color distributions may vary with sensor responses and acquisition conditions, we use this descriptor, together with structure and texture, as an auxiliary statistical target at the scene level rather than as an invariant cross-view measurement.}

\subsubsection{Geometric Structure Mining}
Geometric structure reflects the macroscopic spatial layout and orientation of major scene elements (e.g., road extensions, building alignments). To capture this macroscopic structural information and guide the model to focus on dominant orientations, we leverage the histogram of gradient orientations \citep{struc}. First, we compute the gradient magnitude $M(x,y)$ and orientation $\theta(x,y)\in[0,\pi]$. To mitigate background noise interference and focus on salient structures, an adaptive masking mechanism is introduced. This mechanism filters out low-frequency flat regions via gradient-based relative thresholding, defining the valid structure set $\Omega$:
\begin{equation}
    \Omega=\{(x,y) \mid M(x,y)>\tau_{rel}\cdot \max_{i,j}M(i,j)\}
    \label{eq:omega}
\end{equation}
where $\tau_{rel}$ denotes the suppression coefficient and is set to 0.15. Subsequently, we construct the geometric structure signature $f_{struc}$ by computing the histogram of normalized orientations within $\Omega$. This feature effectively encodes the dominant directionality of the scene, remaining invariant to specific spatial translations.

\subsubsection{Surface Texture Mining}
Surface texture describes the roughness and microscopic complexity of surface materials. To characterize these high-frequency details, we employ the histogram of Laplacian energy distribution \citep{texture}. First, the second-order Laplacian operator $\nabla^{2}$ is applied to extract the high-frequency response energy $E(x,y)=|\nabla^{2}I(x,y)|$. Given that the texture energy of natural scenes typically exhibits a heavy-tailed distribution, direct histogramming would result in excessive clustering in low-energy regions. Therefore, a logarithmic transformation is applied to rectify the distributional skewness, enhancing the separability of high-energy texture details:
\begin{equation}
    E_{log}(x,y)=\log(1+E(x,y))
    \label{eq:log_energy}
\end{equation}
Next, we map the energy to the $[0,1]$ interval via dynamic maximum normalization and compute its distribution histogram as the surface texture signature $f_{tex}$. This signature effectively distinguishes diverse surface material properties by capturing the statistical shape of the energy distribution, thereby enabling the explicit differentiation between sparse strong edges and dense weak textures, {\color{black}complementing color and structure by capturing high-frequency surface roughness and local complexity.} {\color{black}The structure descriptor $f_{struc}$ summarizes the orientation distribution of salient first-order gradients to characterize macroscopic spatial organization, whereas the texture descriptor $f_{tex}$ models the distribution of second-order Laplacian energy to capture local high-frequency surface variation. Together, they characterize scene geometry and surface appearance from different statistical perspectives.}

Ultimately, by concatenating these three physical components, we construct the {\color{black}comprehensive statistical descriptor} $P_{sig}=[f_{color}||f_{struc}||f_{tex}]$, providing a {\color{black}robust statistical cue for scene-level cross-modal alignment}.

\subsection{Multi-modal Feature Encoding and Physical Semantic Projection}
\label{sec:method_encoding}

To establish a robust correspondence between ground-level descriptions and satellite imagery, our PLANET framework operates on a dual-stream architecture that harmonizes high-level global semantics with {\color{black}fine-grained visual statistics}.

\subsubsection{Global Dual-Stream Encoding \& Contrastive Alignment}
We employ a dual-stream backbone consisting of an image encoder $\Phi_{img}$ and a text encoder $\Phi_{txt}$ to extract representations from the satellite image $I$ and the description $T$.

In the visual branch, the image encoder $\Phi_{img}$ maps the satellite image into a normalized global embedding vector $v \in \mathbb{R}^D$, which encapsulates the holistic scene context. In the textual branch, for the transformer-based text encoder $\Phi_{txt}$, we extract features at two distinct levels. First, we obtain a global semantic embedding $t \in \mathbb{R}^D$ representing the macro-meaning of the entire description, which is acquired by extracting the final aggregated features of the text sequence.

To ensure global semantic consistency between modalities, we apply the standard InfoNCE loss \citep{infonce} on these global features. For a batch of $N$ image-text pairs, the global alignment objective $\mathcal{L}_{itc}$ is formulated as:

\begin{equation}
\mathcal{L}_{itc} = -\frac{1}{N}\sum_{i=1}^{N} \log \frac{\exp(\text{sim}(v_i, t_i) / \tau)}{\sum_{j=1}^{N} \exp(\text{sim}(v_i, t_j) / \tau)}
\label{eq:global_loss}
\end{equation}

\noindent where $\text{sim}(\cdot, \cdot)$ denotes cosine similarity and $\tau$ is a learnable temperature parameter {\color{black}for $\mathcal{L}_{itc}$}. While $\mathcal{L}_{itc}$ aligns the general semantic space, it often fails to capture specific physical attributes, such as color distribution or geometric structural orientation, due to the heavy information compression in a single global vector \citep{enhancing}.

\subsubsection{Physical Semantic Projection via Learnable Queries}
To explicitly disentangle intrinsic physical properties from the linguistic stream, we introduce a dedicated physical semantic projection module. Unlike standard linear projections that operate on global tokens, our projection mechanism leverages the full sequence of token embeddings $E_{txt} \in \mathbb{R}^{L \times D}$ from the last hidden layer of the text encoder, where $L$ is the sequence length. This sequence preserves fine-grained semantic details and attribute cues for each word, avoiding information loss.

Inspired by the success of querying mechanisms in recent large vision-language models \citep{blip2}, we implement the projection heads using a set of learnable physical queries, denoted as $Q_{color}, Q_{struc}, Q_{tex} \in \mathbb{R}^D$. These queries act as semantic anchors for the projection, optimized end-to-end to dynamically aggregate attribute-specific information from the sequence $E_{txt}$. Following the standard scaled dot-product attention mechanism \citep{attention}, for each physical dimension $k \in \{color, struc, tex\}$, the projected physical descriptors $\hat{t}_k$ are computed as:
\begin{equation}
\begin{aligned}
A_k &= \text{Softmax}\left( \frac{Q_k (E_{txt} W_K)^T}{\sqrt{D}} \right), \\
\hat{t}_k &= {\color{black}\operatorname{Proj}_k\!\left[\text{LayerNorm}\left( A_k (E_{txt} W_V) \right)\right]}
\end{aligned}
\label{eq:physical_projection}
\end{equation}

\noindent where $A_k \in \mathbb{R}^{1 \times L}$ serves as the projection attention map, highlighting tokens relevant to the specific physical attribute. For instance, the color query vector automatically attends to color-descriptive terms, while the transformation matrices $W_K$ and $W_V$ handle the feature space mapping. {\color{black}$\operatorname{Proj}_k$ is a branch-specific linear projection that maps the attended text descriptor to the dimension of its paired histogram descriptor.} Instead of compressing the sentence into a highly entangled global representation, this query-driven attention mechanism explicitly deconstructs the holistic semantics into a fine-grained form that is dynamically queryable by distinct physical attributes. Consequently, this design encourages the projected descriptors $\hat{t}_k$ to align with {\color{black}the corresponding image statistical descriptors}, facilitating subsequent physical consistency learning.


\subsection{Consistency Contrastive Learning}
\label{sec:consistency_learning}

Following the physical semantic projection, the core challenge of PLANET is to ensure that the projected descriptors $\hat{t}_k$ align with {\color{black}the statistical descriptors extracted from satellite imagery}. To this end, we propose a \textbf{Consistency Contrastive Learning} paradigm that utilizes {\color{black}these image statistical descriptors} as {\color{black}scene-level auxiliary supervision}.

\subsubsection{Intrinsic Physical Supervision}
As described in Sec.~\ref{sec:mining}, we harvest {\color{black}three complementary statistical descriptors} from the satellite imagery: spectral chromaticity $f_{color}$, geometric structure $f_{struc}$, and surface texture $f_{tex}$. Since our physical semantic projectors (Sec.~\ref{sec:method_encoding}) are explicitly designed to map textual representations into the same metric space as these statistical distributions, we {\color{black}use $f_{color}, f_{struc}, f_{tex}$ as scene-level auxiliary targets for their respective physical dimensions, rather than object-level or pixel-level ground truth.}

\subsubsection{Fine-grained Consistency Objective}
To achieve precise geo-localization, the model must explicitly align the textually projected physical semantics with the deterministic visual statistics of the image. For instance, the color descriptors projected by the color query should statistically match the image's color signature. We formulate this via a physical consistency loss $\mathcal{L}_{phy}$. For a batch of $N$ samples, we compute the contrastive loss across the three physical dimensions:


\begin{equation}
\resizebox{0.88\hsize}{!}{$ 
\mathcal{L}_{phy} = \sum\limits_{k} -\frac{1}{N}\sum\limits_{i=1}^{N} \log \frac{\exp(\text{sim}(\hat{t}_{k,i}, f_{k,i}) / \tau_p)}{\sum_{j=1}^{N} \exp(\text{sim}(\hat{t}_{k,i}, f_{k,j}) / \tau_p)}
$}
\label{eq:physical_loss}
\end{equation}

\noindent where $\hat{t}_{k,i}$ is the projected physical descriptor, $f_{k,i}$ is the mined physical signature of the corresponding image, and $\tau_p$ is a {\color{black}separate learnable temperature parameter}. {\color{black}Here, $k \in \{color, struc, tex\}$ denotes the three physical dimensions, with same-sample descriptors forming positive pairs and different-sample descriptors forming in-batch negative pairs. Together, these terms provide a multi-level characterization of scene appearance, covering chromatic composition, macroscopic spatial organization, and local surface variation.} Unlike the global alignment objective, $\mathcal{L}_{phy}$ forces the text encoder to learn interpretably, ensuring that specific linguistic cues (e.g., ``red roofs'', ``grid layout'') are correctly mapped to their corresponding physical manifestations without semantic drift.

\subsubsection{Joint Optimization Objective}
The PLANET framework is optimized in an end-to-end manner. The total loss function $\mathcal{L}_{total}$ fuses the holistic semantic alignment with the fine-grained physical consistency:

\begin{equation}
\mathcal{L}_{total} = \mathcal{L}_{itc} + \lambda \mathcal{L}_{phy}
\label{eq:total_loss}
\end{equation}

\noindent where $\lambda$ {\color{black}is a shared weighting coefficient for the color, structure, and texture terms aggregated in $\mathcal{L}_{phy}$, controlling their overall contribution relative to the global semantic alignment loss $\mathcal{L}_{itc}$.}

{\color{black}The global InfoNCE objective $\mathcal{L}_{itc}$ constrains pairwise similarities between pooled image and text embeddings, leaving the scene evidence supporting each match underdetermined. In contrast, $\mathcal{L}_{phy}$ aligns branch-specific textual projections with satellite-derived statistical targets, introducing constraints that cannot be inferred from global similarities alone. The joint objective therefore provides hierarchical supervision over global semantics and scene statistics, rather than multiple penalties on a common embedding.}

\begin{table*}[pos=t]
\centering
\caption{Quantitative results of PLANET and existing methods on the CORE dataset. We report the retrieval recall ($R@1$) and localization accuracy ($L@150$) across the World-level and four intercontinental subsets. {\color{black}All methods are fine-tuned on the CORE training split under a unified training protocol.} Bold indicates the best result, and underline indicates the second best result.}
\label{tab:comparison}
\resizebox{\textwidth}{!}{%

\begin{tabular}{lcccccccccc}
\toprule
\multirow{3}{*}{\textbf{Method}} & \multicolumn{2}{c}{\textbf{Global-level}} & \multicolumn{8}{c}{\textbf{Intercontinental-level}} \\
\cmidrule(lr){2-3} \cmidrule(lr){4-11}
 & \multicolumn{2}{c}{All} & \multicolumn{2}{c}{Subset1} & \multicolumn{2}{c}{Subset2} & \multicolumn{2}{c}{Subset3} & \multicolumn{2}{c}{Subset4} \\
\cmidrule(lr){2-3} \cmidrule(lr){4-5} \cmidrule(lr){6-7} \cmidrule(lr){8-9} \cmidrule(lr){10-11}
 & R@1 & L@150 & R@1 & L@150 & R@1 & L@150 & R@1 & L@150 & R@1 & L@150 \\
\midrule
ViLT \citep{vilt} & 23.05 & 26.23 & 24.43 & 26.18 & 22.08 & 25.34 & 25.58 & 30.79 & 19.65 & 25.07 \\
X-VLM \citep{XVLM} & 26.97 & 30.81 & 28.90 & 30.65 & 25.72 & 28.74 & 28.67 & 33.86 & 24.31 & 29.62 \\
X2-VLM \citep{X2VLM} & 34.35 & 38.33 & 35.79 & 38.50 & 33.73 & 36.56 & 36.12 & 41.25 & 31.43 & 36.61 \\
SigLIP-B/16 \citep{Siglip} & 30.10 & 34.08 & 29.13 & 31.87 & 30.87 & 33.92 & 31.23 & 36.18 & 28.98 & 34.20 \\
SigLIP-SO400M \citep{Siglip} & 43.27 & 47.48 & 45.03 & 47.74 & 48.26 & 51.34 & 40.71 & 46.15 & 39.14 & 44.69 \\
EVA2-CLIP-B/16 \citep{EVAClip} & 41.02 & 44.17 & 42.18 & 44.74 & 43.77 & 46.96 & 37.32 & 42.37 & 37.39 & 42.73 \\
EVA2-CLIP-L/14 \citep{EVAClip} & 45.38 & 49.56 & 47.59 & 50.40 & 49.44 & 52.52 & 41.03 & 46.31 & 43.98 & 49.41 \\
CLIP-B/16 \citep{CLIP} & 39.67 & 42.69 & 40.91 & 43.02 & 42.36 & 45.37 & 36.09 & 40.73 & 35.84 & 41.37 \\
CLIP-L/14 \citep{CLIP} & 44.63 & 48.92 & 45.75 & 48.42 & 49.19 & 53.14 & 41.48 & 46.56 & 42.53 & 47.76 \\
BLIP \citep{blip} & 42.85 & 46.73 & 45.59 & 48.40 & 48.38 & 51.34 & 39.77 & 44.66 & 37.93 & 42.73 \\
RemoteCLIP \citep{remoteclip} & 42.77 & 46.90 & 44.79 & 47.66 & 47.51 & 50.45 & 40.55 & 45.78 & 38.26 & 43.67 \\
{\color{black}GeoRSCLIP \citep{rs5m}} & {\color{black}46.47} & {\color{black}49.63} & {\color{black}47.10} & {\color{black}50.05} & {\color{black}50.56} & {\color{black}53.83} & {\color{black}41.48} & {\color{black}46.23} & {\color{black}40.65} & {\color{black}46.02} \\
{\color{black}SkyCLIP \citep{skyscript}} & {\color{black}48.84} & {\color{black}52.45} & {\color{black}50.88} & {\color{black}54.01} & {\color{black}57.60} & {\color{black}60.45} & {\color{black}45.03} & {\color{black}50.06} & {\color{black}45.05} & {\color{black}50.58} \\
{\color{black}GRAFT \citep{GRAFT}} & {\color{black}44.33} & {\color{black}48.41} & {\color{black}45.90} & {\color{black}48.58} & {\color{black}49.21} & {\color{black}52.38} & {\color{black}40.68} & {\color{black}45.67} & {\color{black}40.70} & {\color{black}46.63} \\
TAG-CLIP \citep{tagclip} & 38.46 & 42.47 & 38.15 & 40.78 & 42.35 & 45.32 & 37.32 & 42.37 & 36.46 & 41.73 \\
Long-CLIP \citep{longclip} & 51.34 & 55.46 & 52.78 & 55.34 & \underline{59.85} & \underline{62.93} & 45.40 & 50.50 & 47.96 & 53.59 \\
CrossText2Loc \citep{CVGText} & \underline{51.92} & \underline{55.88} & \underline{53.12} & \underline{55.88} & 59.36 & 62.37 & \underline{46.97} & \underline{51.27} & \underline{48.71} & \underline{54.61} \\
\midrule
PLANET (Ours) & \textbf{55.84} & \textbf{59.66} & \textbf{57.90} & \textbf{60.72} & \textbf{64.74} & \textbf{67.58} & \textbf{49.81} & \textbf{54.13} & \textbf{51.35} & \textbf{56.65} \\
\bottomrule
\end{tabular}%
}
\end{table*}

\begin{figure*}[pos=t]
\centering
\includegraphics[width=6.4 in]{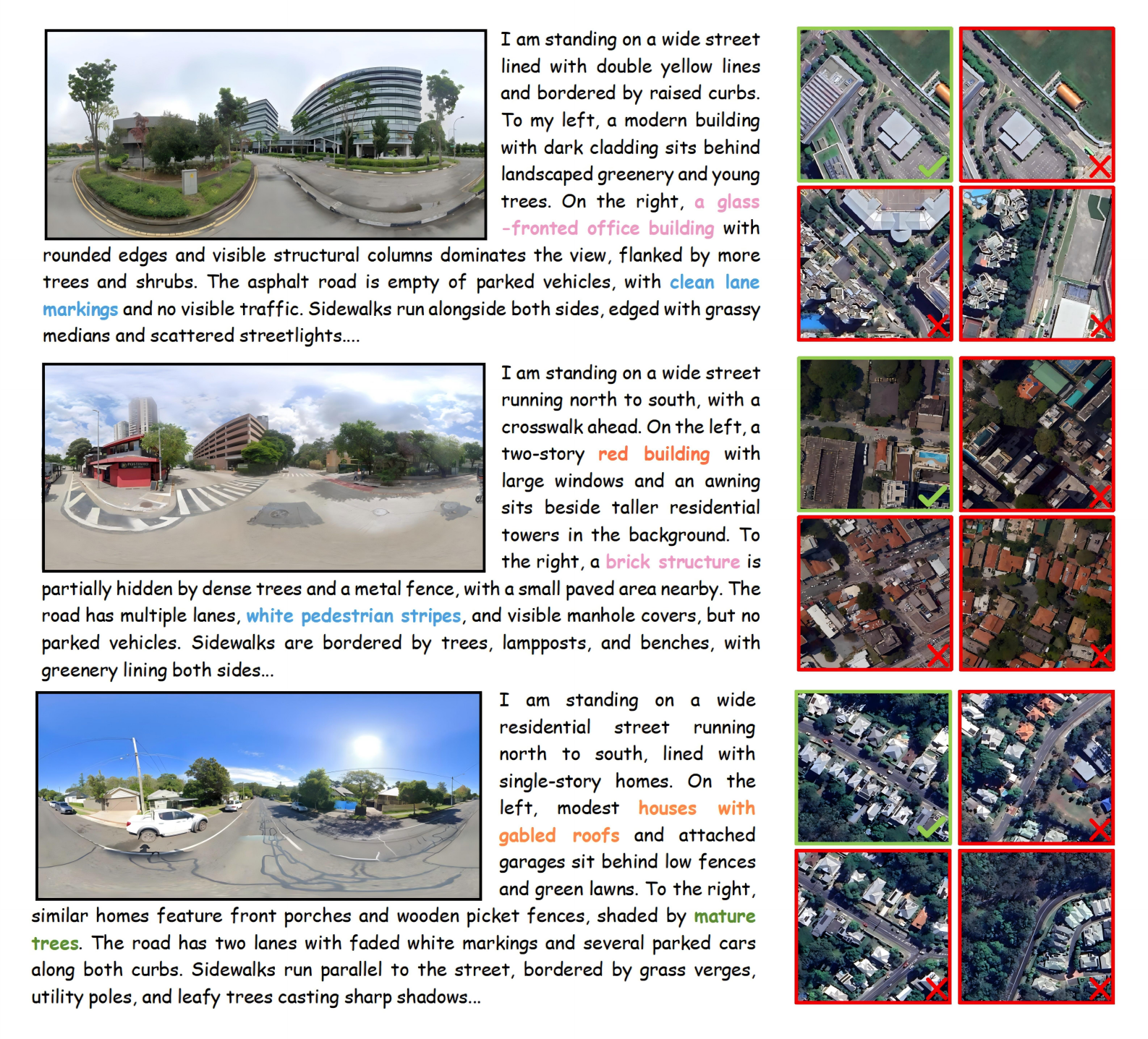}
\caption{The retrieval results of PLANET on the CORE dataset. The left side displays the original street view image along with its corresponding textual description. The right side illustrates the top four retrieval results, where green highlights indicate correct matches and red highlights signify incorrect results.}
\label{VISUAL}
\end{figure*}

\begin{figure*}[pos=t]
\centering
\includegraphics[width=6.4 in]{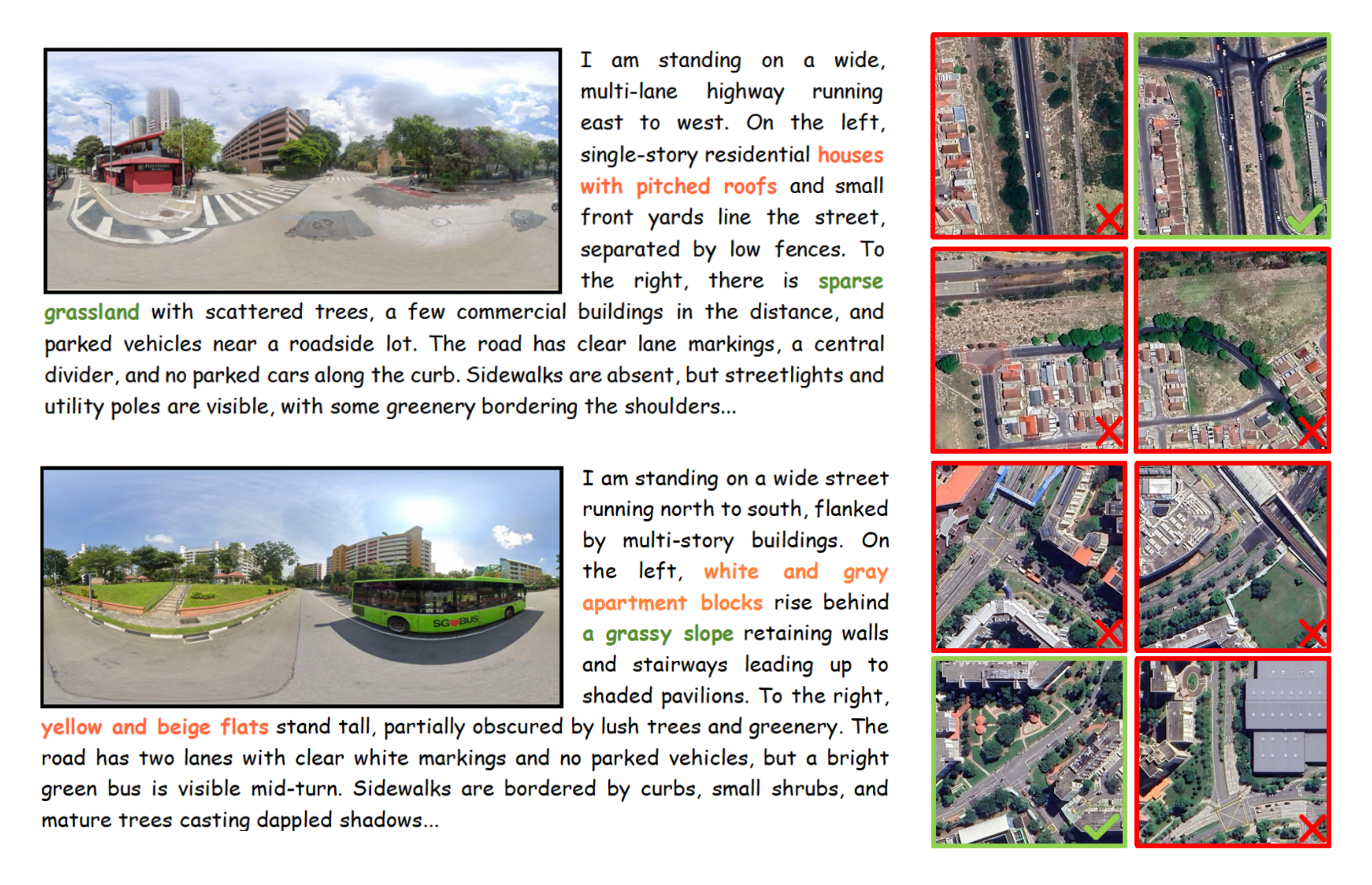}
\caption{{\color{black}Failure cases of PLANET on the CORE dataset. For each query, the right side shows the top four retrieval results in row-major order from Rank-1 to Rank-4. Green boxes denote the ground-truth satellite images, while red boxes denote incorrect retrieved candidates.}}
\label{failurecase}
\end{figure*}

\begin{figure*}[pos=t]
\centering
\includegraphics[width=5.5 in]{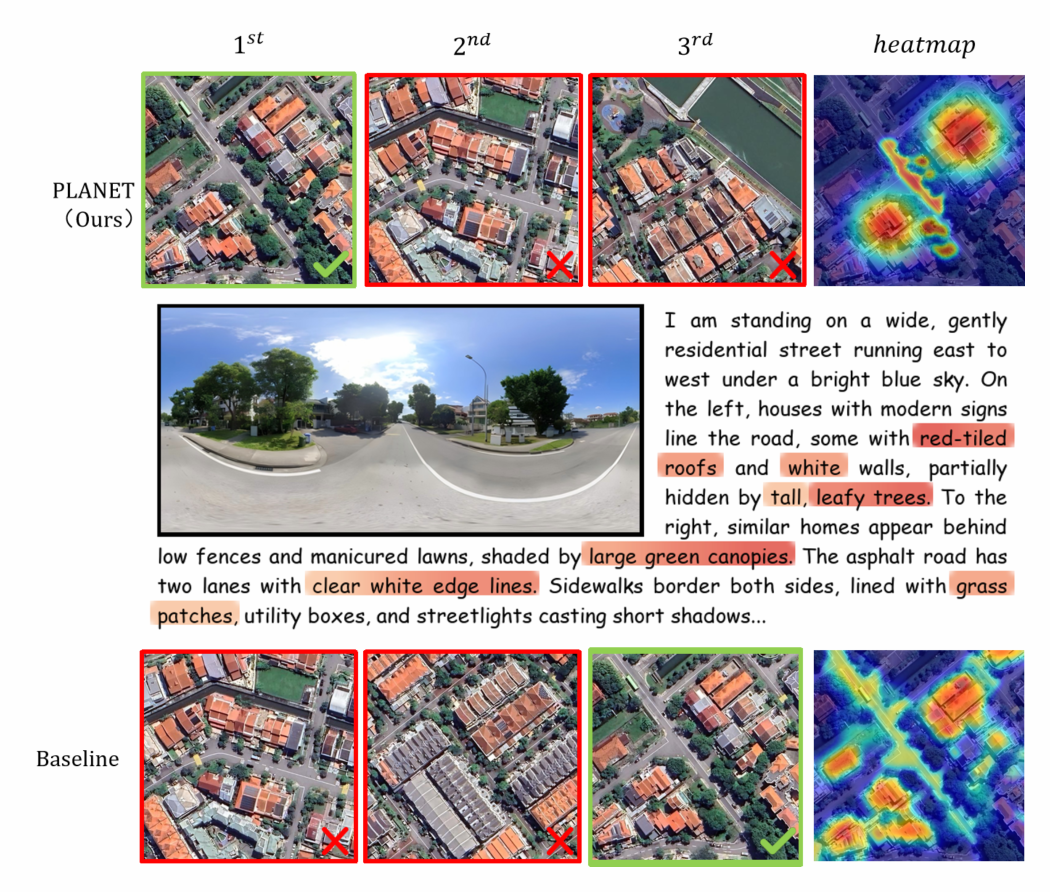}
\caption{Top-3 retrieval results and satellite heatmaps for PLANET and the baseline, {\color{black}together with PLANET's token-level attention visualization of the query text. Green and red boxes denote correct and incorrect retrievals, respectively, and tokens with stronger PLANET attention responses correspond to discriminative physical attributes in the scene.} The introduction of physical consistency constraints enables PLANET to more accurately anchor its attention on crucial geographical regions.}
\label{heatmap}
\end{figure*}

\begin{figure}[pos=t]
\centering
\includegraphics[width=\columnwidth]{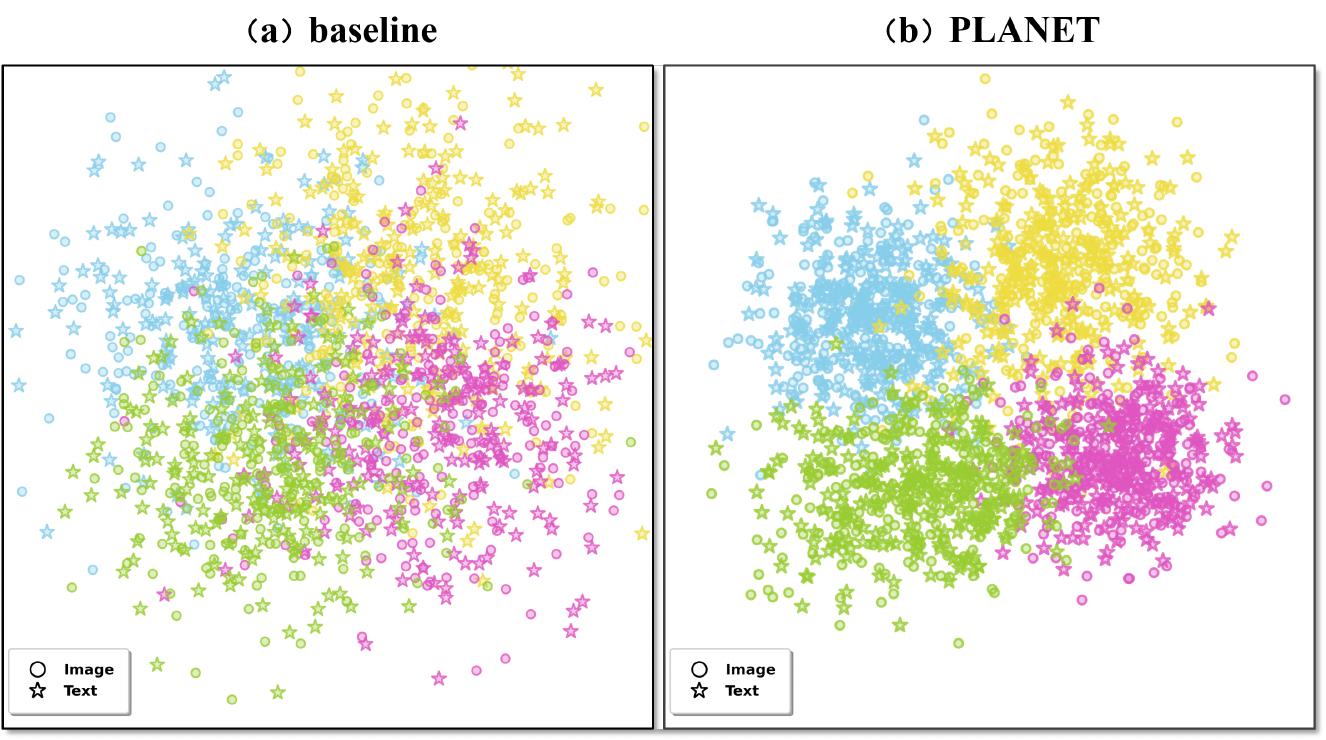}
\caption{UMAP visualization of the feature distribution of the CORE dataset. Compared to the baseline (left), PLANET (right) significantly enhances the discriminative power of features by introducing physical property constraints. Samples from different subsets (distinguished by color) exhibit higher intra-class cohesion and clearer inter-class boundaries in space, while achieving tighter cross-modal alignment between images (circles) and text (stars).}
\label{UMAP}
\end{figure}

\section{Experiments and Analysis}

\subsection{Benchmark}
\label{sec:benchmark}
We establish a benchmark on the CORE dataset for CMGL tasks across two distinct scales: \textbf{World-level} and \textbf{Intercontinental-level}. For a fair comparison, all evaluated methods are fine-tuned on the training split of the corresponding evaluation dataset before testing. The World-level benchmark validates the model's global-scale retrieval capability on the complete dataset, while the Intercontinental-level benchmark evaluates both the domain-specific performance within the four geographical partitions (Sec.~\ref{sec:dataset}) and its out-of-distribution generalization via a cross-regional evaluation paradigm. {\color{black}Compared with city-level CMGL datasets, these settings increase query difficulty by requiring retrieval from a larger and more heterogeneous candidate pool while testing cross-region robustness.} To provide a comprehensive assessment, we adopt retrieval accuracy and localization precision as the two primary evaluation dimensions. For the retrieval dimension, following standard protocols in cross-view geo-localization \citep{CVUSA}, we employ \textbf{$R@1$} {\color{black}(top-1 recall)} as the principal metric. {\color{black}For each textual query, all satellite images in the test gallery are ranked in descending order according to the feature similarity between the textual and visual embeddings. A retrieval is considered successful if the ground-truth satellite image is ranked first, and $R@1$ is calculated as the percentage of successful queries among all textual queries.} For the localization dimension, given the unprecedented geographic breadth and high spatial discreteness of the CORE dataset, we diverge from the conventional $L@50$ standard \citep{text2loc,text2pos} and instead establish \textbf{$L@150$} as the primary metric. This denotes the proportion of queries where the geodesic distance between the predicted result and the actual location is less than 150 meters.

\subsection{Implementation Details}
Regarding data partitioning, we strictly divide each geographical subset of the CORE dataset into training and testing sets at a 9:1 ratio. To mitigate potential spatial leakage, we adopt a spatially disjoint split that prevents geographically adjacent samples with highly similar visual contexts from appearing in both sets. {\color{black}We reserved 10\% of the CORE training split for validation and selected the physical consistency weights solely according to validation $R@1$, then retrained PLANET on the complete training split with the selected weights before test evaluation.} {\color{black}All methods are evaluated under the same test gallery for fair comparison.} {\color{black}For PLANET,} we employ the OpenAI pre-trained CLIP-ViT-L/14@336px \citep{CLIP} model as the backbone for both visual and textual encoders. {\color{black}To reduce the influence of backbone scale, we adopt ViT-L/14 for method configurations in which this backbone is available. Configurations defined with a specific backbone variant, as well as methods based on different visual architectures, retain their designated backbones. Each model uses the input resolution specified by its pretrained configuration.} {\color{black}Following CrossText2Loc \citep{CVGText}, we extend CLIP's text context from 77 to 300 tokens by applying one-dimensional linear interpolation to the original positional embeddings, thereby accommodating long descriptions in both CORE and CVG-Text. All CLIP-based methods use this setting for fair comparison.} All experiments are implemented using the PyTorch framework on two NVIDIA GeForce RTX 4090 GPUs and 48 GB of memory. The models are trained for 40 epochs with a batch size of 32. We utilize the Adam \citep{adam} optimizer with an initial learning rate of $1 \times 10^{-5}$, which is decayed via a cosine annealing schedule. {\color{black}Apart from these configuration-specific input resolutions, all methods use the same data processing pipeline and training configuration, including the number of epochs, batch size, learning rate, optimizer, and learning-rate schedule. No additional data augmentation is applied.}

{\color{black}All physical descriptors are built with 32 histogram bins. The color descriptor concatenates the histograms of three RGB channels and has 96 dimensions, while the structure and texture descriptors each have 32 dimensions. For stable binning, the values used to build the three histograms are mapped to $[0,1]$. Before computing the physical loss, the image histograms and projected text descriptors are both L2-normalized. The color, structure, and texture projectors map textual descriptors to 96, 32, and 32 dimensions, respectively.}

\subsection{Results and Analysis}

 \textit{Comparison With Baselines:} To evaluate the effectiveness of the proposed PLANET framework and provide guidance for future large-scale CMGL research on CORE, we conduct a comprehensive evaluation and analyze the results of {\color{black}17} cross-modal retrieval methods (\citealp{vilt,XVLM,X2VLM,Siglip,EVAClip,CLIP,blip,remoteclip,tagclip,longclip,CVGText}{\color{black}; \citealp{rs5m,skyscript,GRAFT}}). Table~\ref{tab:comparison} provides a comprehensive benchmark on CORE, including detailed quantitative analyses of the ``World-level'' and four ``Intercontinental-level'' subsets. The benchmark methods cover a variety of architectures, including standard vision-language models such as CLIP and SigLIP, as well as specialized spatial retrieval methods such as CrossText2Loc and TAG-CLIP. {\color{black}The newly added remote-sensing baselines further cover large-scale remote-sensing vision-language pretraining and ground-remote alignment.} PLANET significantly outperforms existing methods in all metrics and geographic partitions. Specifically, in highly challenging global benchmarks, PLANET achieves 55.84\% on R@1 and 59.66\% on L@150, representing significant improvements of 3.92\% and 3.78\% in retrieval accuracy and localization accuracy, respectively, compared to the strong baseline CrossText2Loc. Furthermore, PLANET demonstrates consistent performance improvements across all four intercontinental-level subsets, highlighting its robustness to diverse geographic environments and architectural styles. Compared to existing methods, it maintains an average lead of approximately {\color{black}2.64\%--4.89\%} in R@1 and {\color{black}2.04\%--4.84\%} in L@150 in these regions.

 \textit{Qualitative Results and Visualization:} Fig.~\ref{VISUAL} shows representative cross-modal geo-localization visualizations on the CORE dataset. For each query text, we present the top-4 image retrieval results, with true matches marked with green boxes. The results demonstrate that PLANET effectively mines fine-grained physical cues, such as structure, color, and texture, implicit in the query text, thereby accurately retrieving the correct matched satellite images from the candidate lists in various complex scenarios. This demonstrates that PLANET can eliminate intermodal gaps, thus achieving robust cross-modal semantic alignment.
{\color{black}We further analyze representative failure cases in Fig.~\ref{failurecase}.} {\color{black}In the first case, the top-ranked distractor shares a similar road orientation and residential--grassland composition with the ground-truth scene, making the available scene-level cues ambiguous. In the second case, the query emphasizes a temporary bus and building-facade colors that are difficult to observe from the satellite view. These cases indicate that PLANET remains challenged when the transferable evidence shared across views is ambiguous or weak.}

\subsection{Ablation Studies}
\subsubsection{Effectiveness of Physical Consistency Mechanism}
In the methodology phase, we establish a holistic physical consistency mechanism for PLANET, systematically integrating intrinsic physical signature mining, physical semantic projection, and consistency contrastive learning. To validate the effectiveness of this mechanism, we conduct extensive ablation studies across the four intercontinental subsets of the CORE dataset. Specifically, we incrementally introduce individual physical priors into the baseline and compare the results against the fully integrated architecture to assess their respective contributions. As illustrated in Table~\ref{tab:ablation}, the introduction of any single physical constraint ($f_{color}$, $f_{struc}$, or $f_{tex}$) in the baseline consistently enhances performance across all continental subsets. This finding indicates that even single-dimensional physical cues can effectively mitigate the limitations of global semantic features in fine-grained scene differentiation. Quantitatively, individual constraints produce an average improvement in R@1 of 1.87\%--2.00\% over the baseline in diverse regions. {\color{black}Pairwise combinations consistently improve upon the corresponding single-branch variants. In particular, combining structure and texture outperforms either individual branch in both R@1 and L@150 across all four subsets, indicating that macroscopic directional layout and local high-frequency surface variation provide complementary cues for cross-modal retrieval. The complete PLANET model further surpasses the best single-branch variant by 1.40\%--2.16\% in R@1, supporting the joint modeling of the three physical dimensions.} {\color{black}These gains mainly come from using satellite-derived physical statistics to refine textual representations with scene-level targets. Since such targets describe the overall satellite scene rather than local object appearance, they remain robust to cross-view discrepancies and enhance cross-modal discriminability.}

{\color{black}To provide qualitative evidence for the physical consistency mechanism, we compare the satellite heatmaps and retrieval rankings of PLANET and the baseline in Fig.~\ref{heatmap}. PLANET ranks the ground-truth image first, whereas the baseline ranks it third. Its token-level responses emphasize discriminative cues including ``red-tiled roofs'', ``large green canopies'', and ``clear white edge lines'', while its satellite heatmap is more concentrated on the corresponding residential region and surrounding road--vegetation configuration. Together with the branch ablations in Table~\ref{tab:ablation}, these observations suggest that physical consistency supervision helps reduce ambiguity among visually similar candidates.}

\begin{table*}[pos=t]
\centering
\caption{\textbf{Ablation study on the physical consistency mechanism.} We investigate the impact of explicit physical constraints by comparing our PLANET variants against the baseline.}
\label{tab:ablation}
\begin{tabular}{lcccccccc}
\toprule
\multirow{2}{*}{\textbf{Method}} & \multicolumn{2}{c}{\textbf{Subset 1}} & \multicolumn{2}{c}{\textbf{Subset 2}} & \multicolumn{2}{c}{\textbf{Subset 3}} & \multicolumn{2}{c}{\textbf{Subset 4}} \\
\cmidrule(lr){2-3} \cmidrule(lr){4-5} \cmidrule(lr){6-7} \cmidrule(lr){8-9}
 & R@1 & L@150 & R@1 & L@150 & R@1 & L@150 & R@1 & L@150 \\
\midrule
Baseline (w/o Physical) & 53.12 & 55.88 & 59.36 & 62.37 & 46.97 & 51.27 & 48.71 & 54.61 \\
\quad + Color ($f_{color}$) & 55.81 & 58.23 & 62.58 & 65.10 & 47.93 & 52.20 & 49.83 & 55.07 \\
\quad + Structure ($f_{struc}$) & 55.78 & 58.16 & 62.32 & 64.98 & 48.03 & 52.24 & 49.52 & 55.12 \\
\quad + Texture ($f_{tex}$) & 55.43 & 58.10 & 62.46 & 65.08 & 48.21 & 52.53 & 49.95 & 55.48 \\
{\color{black}\quad + Color + Structure ($f_{color}+f_{struc}$)} & {\color{black}56.64} & {\color{black}59.43} & {\color{black}63.98} & {\color{black}66.77} & {\color{black}48.86} & {\color{black}53.21} & {\color{black}50.50} & {\color{black}56.03} \\
{\color{black}\quad + Color + Texture ($f_{color}+f_{tex}$)} & {\color{black}56.26} & {\color{black}59.05} & {\color{black}63.56} & {\color{black}66.36} & {\color{black}49.15} & {\color{black}53.55} & {\color{black}50.82} & {\color{black}56.39} \\
{\color{black}\quad + Structure + Texture ($f_{struc}+f_{tex}$)} & {\color{black}56.19} & {\color{black}59.09} & {\color{black}63.38} & {\color{black}66.19} & {\color{black}48.92} & {\color{black}53.37} & {\color{black}50.67} & {\color{black}56.12} \\
\midrule
\textbf{PLANET (Full)} & \textbf{57.90} & \textbf{60.72} & \textbf{64.74} & \textbf{67.58} & \textbf{49.81} & \textbf{54.13} & \textbf{51.35} & \textbf{56.65} \\
\bottomrule
\end{tabular}%
\end{table*}

\subsubsection{Ablation on Weighting Coefficients of Physical Constraints}

{\color{black}In the standard PLANET configuration, the three physical branches share $\lambda=1/3$. To evaluate the overall strength of physical supervision and the relative contribution of each branch, we untie this shared coefficient only for the ablation study and denote the color, structure, and texture coefficients as $\lambda_c$, $\lambda_s$, and $\lambda_t$, respectively. Their sum controls the overall physical consistency weight, while their proportions determine the relative branch weights.}

{\color{black}Since the coefficient of the global contrastive loss is fixed at 1, we treated 1.0 as the reference value for the total physical consistency weight. With the branch ratio fixed at 1:1:1, we evaluated total weights of 0.3, 0.6, 1.0, 1.5, and 3.0 to cover different levels of physical supervision below and above this reference.}

\begin{table}[pos=htbp]
  \centering
  \caption{{\color{black}Ablation study on the weighting coefficients of the physical consistency constraints across four subsets of the CORE dataset. We report the R@1 (\%) metric.}}
  \label{tab:weight_ablation_core}
  \resizebox{\columnwidth}{!}{
    \begin{tabular}{l ccc c cccc}
      \toprule
      \multirow{2}{*}{\textbf{Variant}} & \multicolumn{3}{c}{\textbf{Weights}} & \multirow{2}{*}{\textbf{$\sum$}} & \multicolumn{4}{c}{\textbf{R@1 (\%) on CORE Subsets}} \\
      \cmidrule(lr){2-4} \cmidrule(lr){6-9}
      & $\lambda_{c}$ & $\lambda_{s}$ & $\lambda_{t}$ & & Sub 1 & Sub 2 & Sub 3 & Sub 4 \\
      \midrule
      \multicolumn{9}{l}{\textit{(a) Ablation on Overall Weight (Fixed 1:1:1 Ratio)}} \\
      Under-regularized & 0.1 & 0.1 & 0.1 & 0.3 & 54.12 {\color{black}$\pm$ 0.07} & 60.83 {\color{black}$\pm$ 0.08} & 47.58 {\color{black}$\pm$ 0.12} & 49.97 {\color{black}$\pm$ 0.04} \\
      Under-regularized & 0.2 & 0.2 & 0.2 & 0.6 & 56.43 {\color{black}$\pm$ 0.06} & 63.15 {\color{black}$\pm$ 0.13} & 48.76 {\color{black}$\pm$ 0.08} & 50.66 {\color{black}$\pm$ 0.07} \\
      Over-regularized  & 0.5 & 0.5 & 0.5 & 1.5 & 57.07 {\color{black}$\pm$ 0.13} & 63.88 {\color{black}$\pm$ 0.11} & 49.26 {\color{black}$\pm$ 0.10} & 50.92 {\color{black}$\pm$ 0.12} \\
      Over-regularized  & 1.0 & 1.0 & 1.0 & 3.0 & 54.82 {\color{black}$\pm$ 0.15} & 60.75 {\color{black}$\pm$ 0.17} & 48.04 {\color{black}$\pm$ 0.15} & 50.23 {\color{black}$\pm$ 0.14} \\
      \midrule
      \multicolumn{9}{l}{\textit{(b) Ablation on Internal Proportions (Fixed $\sum = 1.0$)}} \\
      Biased (Color) & 0.6 & 0.2 & 0.2 & 1.0 & 57.22 {\color{black}$\pm$ 0.08} & 63.85 {\color{black}$\pm$ 0.10} & 49.36 {\color{black}$\pm$ 0.09} & 51.02 {\color{black}$\pm$ 0.12} \\
      Biased (Struc) & 0.2 & 0.6 & 0.2 & 1.0 & 57.08 {\color{black}$\pm$ 0.05} & 63.69 {\color{black}$\pm$ 0.13} & 49.42 {\color{black}$\pm$ 0.12} & 50.88 {\color{black}$\pm$ 0.11} \\
      Biased (Tex) & 0.2 & 0.2 & 0.6 & 1.0 & 56.87 {\color{black}$\pm$ 0.10} & 63.74 {\color{black}$\pm$ 0.08} & 49.49 {\color{black}$\pm$ 0.07} & 51.20 {\color{black}$\pm$ 0.12} \\
      {\color{black}Color+Structure} & {\color{black}0.4} & {\color{black}0.4} & {\color{black}0.2} & {\color{black}1.0} & {\color{black}57.43 $\pm$ 0.11} & {\color{black}64.13 $\pm$ 0.05} & {\color{black}49.43 $\pm$ 0.06} & {\color{black}51.18 $\pm$ 0.06} \\
      {\color{black}Color+Texture} & {\color{black}0.4} & {\color{black}0.2} & {\color{black}0.4} & {\color{black}1.0} & {\color{black}57.36 $\pm$ 0.05} & {\color{black}63.98 $\pm$ 0.09} & {\color{black}49.50 $\pm$ 0.07} & {\color{black}51.23 $\pm$ 0.04} \\
      {\color{black}Structure+Texture} & {\color{black}0.2} & {\color{black}0.4} & {\color{black}0.4} & {\color{black}1.0} & {\color{black}57.28 $\pm$ 0.08} & {\color{black}63.87 $\pm$ 0.10} & {\color{black}49.46 $\pm$ 0.04} & {\color{black}51.21 $\pm$ 0.02} \\
      \midrule
      \textbf{PLANET (Ours)} & \textbf{1/3} & \textbf{1/3} & \textbf{1/3} & \textbf{1.0} & \textbf{57.90 {\color{black}$\pm$ 0.04}} & \textbf{64.74 {\color{black}$\pm$ 0.06}} & \textbf{49.81 {\color{black}$\pm$ 0.03}} & \textbf{51.35 {\color{black}$\pm$ 0.03}} \\
      \bottomrule
    \end{tabular}
  }
\end{table}

First, we evaluate the impact of the overall physical consistency weight by varying the total sum from 0.3 to 3.0 while keeping the internal ratio fixed at 1:1:1. The results in Table~\ref{tab:weight_ablation_core} show that setting the total weight to 1.0 achieves the highest retrieval accuracy across all subsets. This confirms that this scaling factor best balances the physical consistency loss with the cross-modal contrastive loss.

Second, with the total weight fixed at 1.0, we test different internal ratio allocations for the three physical dimensions: color, structure, and texture. As shown in Table~\ref{tab:weight_ablation_core}, {\color{black}the equal-weight configuration achieves the best performance among both single-cue and pair-cue biased settings across all subsets, with small standard deviations across random seeds. A query-level paired randomization test \citep{smucker2007comparison} further verifies that the improvement over the strongest biased setting is statistically significant ($p<0.05$). This indicates that balanced supervision from the three physical cues is more effective than favoring one or two specific cues.}

\subsubsection{{\color{black}Ablation on Histogram Bin Number}}

{\color{black}To evaluate the sensitivity of PLANET to histogram discretization, we vary the number of bins used to construct the physical descriptors while keeping all other settings unchanged. The results are reported in Table~\ref{tab:bin_ablation}.}

\begin{table}[pos=t]
\centering
\caption{{\color{black}Ablation study on the histogram bin number of physical descriptors. We report R@1 (\%) and L@150 (\%) across four subsets of the CORE dataset.}}
\label{tab:bin_ablation}
\resizebox{\columnwidth}{!}{
\begin{tabular}{lcccccccc}
\toprule
\multirow{2}{*}{{\color{black}\textbf{Bins}}} & \multicolumn{2}{c}{{\color{black}\textbf{Subset 1}}} & \multicolumn{2}{c}{{\color{black}\textbf{Subset 2}}} & \multicolumn{2}{c}{{\color{black}\textbf{Subset 3}}} & \multicolumn{2}{c}{{\color{black}\textbf{Subset 4}}} \\
\cmidrule(lr){2-3} \cmidrule(lr){4-5} \cmidrule(lr){6-7} \cmidrule(lr){8-9}
 & {\color{black}R@1} & {\color{black}L@150} & {\color{black}R@1} & {\color{black}L@150} & {\color{black}R@1} & {\color{black}L@150} & {\color{black}R@1} & {\color{black}L@150} \\
\midrule
{\color{black}8} & {\color{black}57.08} & {\color{black}59.76} & {\color{black}63.88} & {\color{black}66.57} & {\color{black}48.95} & {\color{black}53.37} & {\color{black}50.39} & {\color{black}55.88} \\
{\color{black}16} & {\color{black}57.43} & {\color{black}60.27} & {\color{black}64.32} & {\color{black}67.13} & {\color{black}49.28} & {\color{black}53.64} & {\color{black}50.95} & {\color{black}56.13} \\
{\color{black}\textbf{32}} & {\color{black}\textbf{57.90}} & {\color{black}\textbf{60.72}} & {\color{black}\textbf{64.74}} & {\color{black}\textbf{67.58}} & {\color{black}\textbf{49.81}} & {\color{black}\textbf{54.13}} & {\color{black}\textbf{51.35}} & {\color{black}\textbf{56.65}} \\
{\color{black}64} & {\color{black}57.52} & {\color{black}60.33} & {\color{black}64.38} & {\color{black}67.09} & {\color{black}49.21} & {\color{black}53.58} & {\color{black}51.02} & {\color{black}56.27} \\
{\color{black}128} & {\color{black}57.17} & {\color{black}59.91} & {\color{black}64.03} & {\color{black}66.75} & {\color{black}48.88} & {\color{black}53.26} & {\color{black}50.62} & {\color{black}55.95} \\
\bottomrule
\end{tabular}
}
\end{table}

{\color{black}Table~\ref{tab:bin_ablation} shows that using 32 bins achieves the best results across all subsets and metrics, suggesting a good balance between statistical granularity and distribution stability. We use the same bin number for all three physical branches to keep the discretization controlled and comparable. Although the color descriptor concatenates three RGB channel histograms, independent normalization and branch-wise loss computation make the three branches comparable during training. The relatively stable performance under different bin numbers further indicates that the gains mainly come from incorporating statistical cues from satellite images, rather than from a particular histogram discretization.}

\subsection{{\color{black}Efficiency Analysis with Different Backbones}}

{\color{black}To evaluate the deployment flexibility of PLANET, we compare the original baseline with PLANET using different CLIP backbones on CORE Subset 1. During inference, PLANET relies only on standard text-image feature matching, since the statistical descriptors are used exclusively for training supervision. All efficiency measurements are conducted under the same batch size. As shown in Table~\ref{tab:efficiency}, compared with the baseline using CLIP-ViT-L/14@336px, PLANET with the same backbone increases the parameters by only 3.67M, which is less than 1\% of the baseline. The FLOPs increase is also marginal, and the retrieval time remains nearly identical to the baseline, while R@1 and L@150 improve by 4.78\% and 4.84\%, respectively. These results show that the proposed modules bring clear accuracy gains with only marginal additional parameters and computation. The results with CLIP-ViT-B/32 and CLIP-ViT-B/16 further show that PLANET supports lighter backbones, but with a trade-off in retrieval accuracy.}

\begin{table*}[pos=t]
\centering
\caption{{\color{black}Efficiency and accuracy comparison on CORE Subset 1. The baseline uses the original CLIP-ViT-L/14@336px backbone, while PLANET is evaluated with different CLIP backbones. Retrieval time denotes the average retrieval time for a single query.}}
\label{tab:efficiency}
\resizebox{0.85\textwidth}{!}{
\begin{tabular}{llcccccc}
\toprule
{\color{black}\textbf{Method}} & {\color{black}\textbf{Backbone}} & {\color{black}\textbf{Params (M)}} & {\color{black}\textbf{FLOPs (G)}} & {\color{black}\textbf{Memory (GB)}} & {\color{black}\textbf{Retrieval Time (ms)}} & {\color{black}\textbf{R@1}} & {\color{black}\textbf{L@150}} \\
\midrule
{\color{black}Baseline} & {\color{black}CLIP-ViT-L/14@336px} & {\color{black}428.12} & {\color{black}401.03} & {\color{black}30.45} & {\color{black}49.97} & {\color{black}53.12} & {\color{black}55.88} \\
\midrule
\multirow{3}{*}{{\color{black}PLANET}} 
& {\color{black}CLIP-ViT-B/32} & {\color{black}155.06} & {\color{black}32.58} & {\color{black}6.43} & {\color{black}31.28} & {\color{black}41.03} & {\color{black}44.22} \\
& {\color{black}CLIP-ViT-B/16} & {\color{black}153.40} & {\color{black}57.57} & {\color{black}8.40} & {\color{black}31.93} & {\color{black}44.49} & {\color{black}47.22} \\
& {\color{black}CLIP-ViT-L/14@336px} & {\color{black}431.79} & {\color{black}403.16} & {\color{black}30.46} & {\color{black}50.08} & {\color{black}57.90} & {\color{black}60.72} \\
\bottomrule
\end{tabular}
}
\end{table*}

\subsection{{\color{black}Textual Localization Cue Analysis}}

{\color{black}To examine which parts of the query text support localization, we conduct a cue ablation experiment on CORE Subset 1 during testing. Starting from the full query, we remove entity words, all attribute words, or one of the three physical cue groups including color, structure, and texture. We also test an entity only setting that retains only entity words. The trained model and retrieval gallery remain unchanged. As shown in Table~\ref{tab:text_cue_removal}, removing entities or all attributes leads to large drops, indicating that object nouns provide scene anchors and attributes supply detailed discrimination. The entity only setting drops to 10.93\% R@1 and 12.52\% L@150, showing that highly compact entity concepts alone cannot support reliable localization. Among the physical cue groups, removing structure causes the largest drop, followed by color and texture. This suggests that road and layout cues are especially important for matching ground descriptions with overhead satellite scenes, while color and texture provide complementary evidence.}

\begin{table}[pos=t]
\centering
\caption{{\color{black}\textbf{Text cue ablation study on CORE Subset 1.} We remove one cue group from the query during testing, while the Entity only setting retains only entity words. The trained model and retrieval gallery remain unchanged.}}
\label{tab:text_cue_removal}
\resizebox{\columnwidth}{!}{%
\begin{tabular}{lcccc}
\toprule
{\color{black}\textbf{Setting}} & {\color{black}\textbf{R@1}} & {\color{black}\textbf{L@150}} & {\color{black}\textbf{R@1 Drop}} & {\color{black}\textbf{L@150 Drop}} \\
\midrule
{\color{black}Full text} & {\color{black}\textbf{57.90}} & {\color{black}\textbf{60.72}} & {\color{black}--} & {\color{black}--} \\
{\color{black}Entity only} & {\color{black}10.93} & {\color{black}12.52} & {\color{black}46.97} & {\color{black}48.20} \\
{\color{black}w/o Entity} & {\color{black}36.20} & {\color{black}38.86} & {\color{black}21.70} & {\color{black}21.86} \\
{\color{black}w/o Attribute} & {\color{black}34.66} & {\color{black}36.98} & {\color{black}23.24} & {\color{black}23.74} \\
{\color{black}w/o Color} & {\color{black}49.30} & {\color{black}52.05} & {\color{black}8.60} & {\color{black}8.67} \\
{\color{black}w/o Structure} & {\color{black}45.24} & {\color{black}47.74} & {\color{black}12.66} & {\color{black}12.98} \\
{\color{black}w/o Texture} & {\color{black}52.55} & {\color{black}55.34} & {\color{black}5.35} & {\color{black}5.38} \\
\bottomrule
\end{tabular}}
\end{table}

\begin{table}[pos=t]
\centering

\caption{Quantitative comparison of out-of-distribution generalization on CORE dataset.}
\label{tab:generalization}
\resizebox{\columnwidth}{!}{
\begin{tabular}{lcccccc}
\toprule
\multirow{2}{*}{\textbf{Method}} & \multicolumn{2}{c}{Subset1 $\rightarrow$ 4} & \multicolumn{2}{c}{Subset2 $\rightarrow$ 4} & \multicolumn{2}{c}{Subset3 $\rightarrow$ 4} \\
\cmidrule(lr){2-3} \cmidrule(lr){4-5} \cmidrule(lr){6-7}
 & R@1 & L@150 & R@1 & L@150 & R@1 & L@150 \\
\midrule
TAG-CLIP & 12.22 & 14.13 & 11.83 & 13.79 & 14.79 & 17.06 \\
BLIP & 15.24 & 17.93 & 14.17 & 16.58 & 16.84 & 19.35 \\
CLIP-L/14 & 15.58 & 18.25 & 14.34 & 16.62 & 17.22 & 19.83 \\
Long-CLIP & 21.68 & 25.07 & 19.75 & 22.86 & 22.96 & 26.10 \\
CrossText2Loc & \underline{23.13} & \underline{26.97} & \underline{21.06} & \underline{24.64} & \underline{24.75} & \underline{28.12} \\
\midrule
PLANET (Ours) & \textbf{28.32} & \textbf{32.54} & \textbf{26.80} & \textbf{30.88} & \textbf{30.96} & \textbf{35.01} \\
\bottomrule
\end{tabular}
}
\end{table}

\begin{table}[pos=t]
\centering
\caption{Quantitative comparison of different methods on the CVG-Text dataset.}
\label{tab:comparison_simplified}

\resizebox{\columnwidth}{!}{%

\begin{tabular}{lcccccc}
\toprule
\multirow{2}{*}{\textbf{Method}} & \multicolumn{2}{c}{New York} & \multicolumn{2}{c}{Brisbane} & \multicolumn{2}{c}{Tokyo} \\
\cmidrule(lr){2-3} \cmidrule(lr){4-5} \cmidrule(lr){6-7}
 & R@1 & L@50 & R@1 & L@50 & R@1 & L@50 \\
\midrule
ViLT \citep{vilt} & 11.58 & 15.58 & 11.10 & 14.50 & 10.83 & 15.50 \\
X-VLM \citep{XVLM} & 15.74 & 16.86 & 15.67 & 17.60 & 12.46 & 14.34 \\
X2-VLM \citep{X2VLM} & 23.33 & 25.17 & 21.58 & 23.67 & 17.83 & 20.75 \\
SigLIP-B/16 \citep{Siglip} & 19.67 & 21.08 & 19.58 & 22.00 & 15.58 & 17.92 \\

SigLIP-SO400M \citep{Siglip} & 33.50 & 34.83 & 34.25 & 36.83 & 28.42 & 31.50 \\
EVA2-CLIP-B/16 \citep{EVAClip} & 25.17 & 26.58 & 28.42 & 31.75 & 22.50 & 25.25 \\
EVA2-CLIP-L/14 \citep{EVAClip} & 34.08 & 35.67 & 35.67 & 38.00 & 31.00 & 34.08 \\
CLIP-B/16 \citep{CLIP} & 23.86 & 25.03 & 26.91 & 30.08 & 21.12 & 23.58 \\
CLIP-L/14 \citep{CLIP} & 35.08 & 37.08 & 34.08 & 37.25 & 28.08 & 30.50 \\
BLIP \citep{blip} & 34.58 & 37.25 & 34.50 & 38.17 & 29.75 & 33.67 \\
RemoteCLIP \citep{remoteclip} & 33.83 & 36.33 & 32.92 & 34.75 & 26.42 & 29.08 \\
{\color{black}GeoRSCLIP \citep{rs5m}} & {\color{black}36.58} & {\color{black}38.83} & {\color{black}35.75} & {\color{black}38.25} & {\color{black}29.92} & {\color{black}32.83} \\
{\color{black}SkyCLIP \citep{skyscript}} & {\color{black}39.83} & {\color{black}42.17} & {\color{black}38.92} & {\color{black}41.75} & {\color{black}32.83} & {\color{black}35.50} \\
{\color{black}GRAFT \citep{GRAFT}} & {\color{black}34.75} & {\color{black}37.00} & {\color{black}33.58} & {\color{black}36.17} & {\color{black}28.58} & {\color{black}31.42} \\
TAG-CLIP \citep{tagclip} & 31.17 & 33.25 & 29.83 & 31.75 & 24.33 & 28.08 \\
Long-CLIP \citep{longclip} & 46.42 & 48.25 & 46.75 & 48.17 & 33.83 & 35.50 \\
CrossText2Loc \citep{CVGText} & \underline{50.33} & \underline{53.08} & \underline{47.58} & \underline{51.80} & \underline{41.75} & \underline{43.86} \\
\midrule
PLANET (Ours) & \textbf{52.83} & \textbf{55.58} & \textbf{49.83} & \textbf{53.67} & \textbf{43.42} & \textbf{46.42} \\
\bottomrule
\end{tabular}
}
\end{table}

\subsection{Cross-Subset Generalization Experiments}

The heterogeneous nature of the CORE dataset, characterized by architectural styles and natural environments, inherently introduces significant domain shifts. To visualize how models handle this diversity under standard supervision, we project the feature embeddings of a model trained on the full dataset using UMAP technique \citep{umap}. As illustrated in Fig.~\ref{UMAP}, PLANET exhibits significantly tighter intra-class compactness and superior cross-modal alignment compared to baseline, where image and text features are densely coupled. While this highly discriminative latent structure correlates with our superior localization accuracy on the standard test set, it raises a critical question: Does this tight coupling stem from learning {\color{black}transferable scene-level cues}, or merely from overfitting to domain-specific statistics?

To rigorously decouple these factors and evaluate the model’s adaptability to unseen geographic environments, we conduct cross-domain generalization experiments using Subset 4 as the target test domain. This selection is motivated by the fact that the African and South American regions represented in Subset 4 diverge significantly from the modern urban landscapes dominant in other training subsets. They are characterized by unique physical attributes, such as unpaved road networks, specific vegetation patterns, and informal architectural styles. These distinctions constitute a substantial distribution shift, posing a severe challenge to the model's ability to extract {\color{black}features that transfer across regions}. Specifically, we trained the model separately on Subsets 1, 2, and 3 and subsequently validated it on the test set of Subset 4.

Table~\ref{tab:generalization} presents the quantitative results of the cross-domain generalization experiments. Due to the distinct natural landscapes and architectural styles of the target domain compared to the training domains, all baseline models suffer severe performance degradation in the unseen environment, underscoring the formidable challenge posed by domain shifts. However, across all cross-domain settings, PLANET exhibits exceptional robustness, comprehensively and significantly outperforming the state-of-the-art methods.
Specifically, in tasks where PLANET migrates from Subset 1, 2, and 3 to Subset 4, its R@1 achieves absolute performance improvements of 5.19\%, 5.74\%, and 6.21\%, respectively, fully demonstrating its outstanding robustness in capturing {\color{black}transferable scene-level visual cues}. This result validates our earlier hypothesis: the highly compact latent space learned by PLANET stems not from overfitting to domain-specific statistics, but from its successful learning of {\color{black}scene-level statistical regularities that generalize across regions}. Driven by explicit physical consistency constraints, the proposed model effectively resists drastic geographic variations, enabling accurate and stable cross-modal localization in highly heterogeneous global scenarios.

\subsection{Results on Existing Dataset}

To further validate the generalization ability and robustness of our proposed PLANET, as well as to evaluate its city-level localization performance, we conduct comparative experiments using the widely recognized CVG-Text dataset \citep{CVGText}. This dataset emphasizes typical modern urban environments, providing a rigorous testing platform to assess whether models can consistently establish fine-grained cross-modal physical alignments within complex, high-density urban layouts.

The quantitative results shown in Table~\ref{tab:comparison_simplified} indicate that PLANET achieves state-of-the-art performance across all three cities, significantly surpassing existing large vision-language models and specialized geo-localization networks. Specifically, in comparison to the strongest baseline, CrossText2Loc, PLANET demonstrates absolute improvements in the R@1 metric of 2.50\%, 2.25\%, and 1.67\% for New York, Brisbane, and Tokyo, respectively, along with consistent gains in the L@50 metric. These findings provide compelling evidence that our proposed physical consistency mechanism not only excels in highly heterogeneous global landscapes but also maintains superior feature discriminability in conventional urban scenarios.

\subsection{{\color{black}Cross-Dataset Generalization Experiments}}

\begin{table}[pos=t]
\centering
\caption{{\color{black}Cross-dataset generalization of PLANET between CORE and CVG-Text. The model is trained on the source dataset and directly evaluated on the target dataset without additional fine-tuning.}}
\label{tab:cross_dataset_generalization}
\resizebox{\columnwidth}{!}{
\begin{tabular}{llccc}
\toprule
{\color{black}\textbf{Training Set}} & {\color{black}\textbf{Test Set}} & {\color{black}\textbf{R@1}} & {\color{black}\textbf{L@50}} & {\color{black}\textbf{L@150}} \\
\midrule
\multirow{3}{*}{{\color{black}CORE}} & {\color{black}CVG-Text New York} & {\color{black}23.17} & {\color{black}25.25} & {\color{black}29.42} \\
 & {\color{black}CVG-Text Brisbane} & {\color{black}26.33} & {\color{black}29.83} & {\color{black}30.50} \\
 & {\color{black}CVG-Text Tokyo} & {\color{black}19.92} & {\color{black}22.75} & {\color{black}23.58} \\
\midrule
\multirow{4}{*}{{\color{black}CVG-Text}} & {\color{black}CORE Subset1} & {\color{black}8.03} & {\color{black}8.85} & {\color{black}10.36} \\
 & {\color{black}CORE Subset2} & {\color{black}9.14} & {\color{black}9.98} & {\color{black}12.57} \\
 & {\color{black}CORE Subset3} & {\color{black}7.75} & {\color{black}8.52} & {\color{black}9.85} \\
 & {\color{black}CORE Subset4} & {\color{black}8.24} & {\color{black}9.33} & {\color{black}10.81} \\
\bottomrule
\end{tabular}
}
\end{table}

{\color{black}Table~\ref{tab:cross_dataset_generalization} shows that direct transfer between CORE and CVG-Text suffers clear performance degradation compared with training and testing on the same dataset, mainly due to differences in text description style, visual content distribution, and geographic coverage. The low scores obtained when training on CVG-Text and testing on CORE reflect the limited coverage of the source dataset when facing a much broader and more diverse target domain. In contrast, the model trained on CORE transfers better to CVG-Text, suggesting that CORE contains more diverse matching patterns and provides more generalizable supervision for global CMGL.}

\subsection{Limitation}
Although our PLANET framework demonstrates exceptional performance and stability in global CMGL, it is still subject to an inherent limitation: the challenge of achieving absolute temporal synchronization between street-view and satellite imagery acquisition. Transient visual elements, such as vehicles and construction sites, introduce unavoidable noise into the cross-modal alignment process. While our intrinsic physical signature mining alleviates this issue to some extent, achieving robust localization in extreme scenarios—such as conflict zones or natural disasters—remains highly challenging. Future research will focus on data acquisition under extreme conditions and time-invariant feature mining to promote feature alignment across different modalities and further improve generalization.

\section{Conclusion}
In this paper, we propose CORE, the first million-scale dataset dedicated to global-scale cross-modal geo-localization (CMGL). This dataset comprises 1,034,786 multi-view images captured from both ground and aerial perspectives across 225 distinct geographic regions, accompanied by textual descriptions for each ground image. The CORE dataset is highly valuable due to its substantial sample size, fine textual granularity, and pronounced geographic heterogeneity. It is anticipated to facilitate a new, challenging, yet significant task: global-scale geo-localization.
Furthermore, we present PLANET, a physical-law-aware consistency contrastive learning paradigm that explicitly aligns {\color{black}statistical descriptors of color, geometric structure, and surface texture} between imagery and text. Utilizing the CORE dataset, we establish a benchmark and empirically validate the effectiveness and generalization capability of the proposed PLANET. In future work, we aim to further enhance the scene diversity of the CORE dataset and investigate the generalizability of {\color{black}mechanisms guided by visual statistics} across a wider array of vision-language understanding tasks.

\vspace{10pt}
\noindent
\textbf{Acknowledgment}
\vspace{10pt}

This work was supported in part by the Xiong'an New Area Science and Technology Innovation Special Program under Grant 2025XAGG0042, in part by the National Natural Science Foundation of China (NSFC) under Grant 42271444, and in part by the Postdoctoral Fellowship Program of CPSF under Grant Number GZB20250065.

\printcredits

\section*{Declaration of competing interest}
The authors declare that they have no known competing financial interests or personal relationships that could have appeared to influence the work reported in this paper.

\section*{Data availability}
The dataset and source code will be made available at https://github.com/YtH0823/CORE.

\makeatletter
\let\CORE@origbibitem\bibitem
\def\CORE@redbibkey#1{\expandafter\def\csname CORE@redbib@#1\endcsname{}}
\CORE@redbibkey{DRL}
\CORE@redbibkey{p2fcn}
\CORE@redbibkey{EAGLe}
\CORE@redbibkey{smucker2007comparison}
\CORE@redbibkey{rs5m}
\CORE@redbibkey{skyscript}
\CORE@redbibkey{GRAFT}
\def\CORE@roundtwobibkey#1{\expandafter\def\csname CORE@roundtwobib@#1\endcsname{}}
\CORE@roundtwobibkey{Zhang_2026_CVPR}
\CORE@roundtwobibkey{zhang2022cvpr}
\CORE@roundtwobibkey{zhang2024tpami}
\CORE@roundtwobibkey{zhang2024ijcv}
\CORE@roundtwobibkey{zhang2024cvpr}
\def\CORE@markbibcolor#1{%
  \color{black}%
  \@ifundefined{CORE@redbib@#1}{}{\color{black}}%
  \@ifundefined{CORE@roundtwobib@#1}{}{\color{black}}%
}
\def\bibitem{%
  \@ifnextchar[{\CORE@bibitem@opt}{\CORE@bibitem@noopt}%
}
\def\CORE@bibitem@opt[#1]#2{%
  \color{black}%
  \CORE@origbibitem[#1]{#2}%
  \CORE@markbibcolor{#2}%
}
\def\CORE@bibitem@noopt#1{%
  \color{black}%
  \CORE@origbibitem{#1}%
  \CORE@markbibcolor{#1}%
}
\makeatother
\bibliographystyle{cas-model2-names}
\bibliography{cas-refs}

\end{document}